\documentclass[amsmath, amssymb, aps, pra, twocolumn, floatfix, nofootinbib, showkeys]{revtex4}

\usepackage[urlcolor=blue, colorlinks=true,citecolor=blue]{hyperref}
\usepackage{graphicx}
\usepackage{dcolumn}
\usepackage{bm}
\usepackage{amsmath}
\usepackage{float}
\usepackage{amssymb}
\usepackage{color}
\usepackage{xcolor}
\usepackage[english]{babel}
\usepackage{braket}
\usepackage{multirow}
\usepackage{booktabs}
\usepackage{algorithm}
\usepackage{algpseudocode}

\begin{document}

\preprint{APS/123-QED}

\title{Hybrid quantum recurrent neural network for remaining useful life prediction of turbofan engines}

\author{Olga Tsurkan}
\author{Aleksandra Konstantinova}
\author{Arsenii Senokosov}
\author{Asel Sagingalieva}
\author{Alexey Melnikov}
\thanks{Corresponding author: \texttt{alexey@melnikov.info}}
\affiliation{Terra Quantum AG, Kornhausstrasse 25, 9000 St.~Gallen, Switzerland}

\begin{abstract}
Accurate remaining useful life (RUL) estimation underpins safe operation and cost-effective maintenance of aerospace propulsion systems. We propose a Hybrid Quantum Recurrent Neural Network (HQRNN) for jet-engine RUL forecasting on the NASA C-MAPSS FD001 benchmark. The HQRNN stacks Quantum Long Short-Term Memory (QLSTM) layers, replacing each LSTM gate's linear transformation with a Quantum Depth-Infused (QDI) circuit; this is followed by classical dense layers. Quantum and hybrid quantum--classical methods for turbofan RUL prediction are still at an early stage. Our study is therefore among the first to evaluate a gate-based QLSTM hybrid at matched parameter counts, comparing it against classical and joint state-of-the-art models on this benchmark and complementing that comparison with a circuit-level analysis of the quantum layer. Encoding the gate signals in a quantum feature space is intended to help the network represent high-frequency degradation patterns with fewer trainable parameters than a matched classical counterpart. The HQRNN improves mean RMSE and mean MAE by about $5\%$ over matched-parameter stacked-LSTM RNNs across 10 random seeds, and attains a test RMSE of $15.46$, outperforming Random Forest, CNN, and MLP baselines. ZX calculus, Fisher information, and Fourier analyses indicate that the QDI circuit is compact, trainable, and expressive. Advanced joint deep-learning models still outperform the stand-alone HQRNN, indicating that quantum-enhanced recurrent modules are best deployed as components within composite prognostics pipelines rather than stand-alone predictors.
\end{abstract}


\maketitle

\begin{figure}[b!]
\fbox{\parbox{\dimexpr\columnwidth-2\fboxsep-2\fboxrule\relax}{%
\footnotesize\raggedright Please check the published version, which includes all the latest additions and corrections: Algorithms 19(8):663, 2026, DOI: \href{https://doi.org/10.3390/a19080663}{10.3390/a19080663}%
}}
\end{figure}

\section{Introduction}\label{sec:introduction}

Accurate estimation of the remaining useful life (RUL) of safety-critical
machinery is a cornerstone of modern reliability and risk-management
strategies \cite{si2011remaining,lee2014prognostics,berghout2022systematic}.
Nowhere is this more evident than in commercial aviation, where the timely
prediction of gas-turbine jet engine failures both prevents catastrophic
breakdowns \cite{zhang2016multiobjective} and streamlines maintenance and
resource allocation \cite{kang2021remaining}. Robust forecasting models enable
operators to defer costly repairs until necessary while avoiding the risks of
overdue overhauls, ultimately reducing downtime and improving fleet
utilisation \cite{huang2019bidirectional,ferreira2022remaining}.

Time-series methods for RUL estimation can be broadly categorised into
statistical and machine-learning approaches. Traditional statistical methods
such as autoregressive, moving-average and ARIMA models excel under
stationarity and linearity assumptions~\cite{box2015time,shumway2000time} but
struggle to capture the non-linear, interdependent patterns characteristic of
real-world aerospace data \cite{zhang2003time}. Within machine learning,
recurrent neural networks (RNNs) and, in particular, Long Short-Term Memory
(LSTM) networks \cite{hochreiter1997long,gers2000learning,graves2013speech,sak2014long}
have become standard tools for capturing temporal dependencies in degradation
signals~\cite{LSTM2024Summary}.

Despite these successes, classical machine-learning strategies can be weakened
by limited or noisy run-to-failure data, high-dimensional feature spaces, and
intricate fault dynamics \cite{Bishop2007Pattern,Goodfellow2016Deep}. These
conditions are routine in industrial prognostics: run-to-failure datasets are
expensive to acquire, often containing only tens or hundreds of full
trajectories, and sensor readings are noisy and correlated. Methods that can
generalise well from limited data are therefore highly desirable.

Quantum computing has recently progressed from theoretical constructs to
initial practical demonstrations, leveraging entanglement and superposition to
enable computational processes beyond conventional
hardware \cite{Nielsen2011Quantum,biamonte2017quantum,montanaro2016quantum,Preskill2018quantumcomputingin}.
Quantum machine learning (QML) emerged at the intersection of these two
fields, with theoretical and empirical evidence that QML models can address
problems characterised by constrained datasets or high-dimensional feature
spaces \cite{rebentrost2014quantum,ciliberto2018quantum,schuld2018supervised,cao2017quantum,Alharbi2024QCNN}.
QML leverages high-dimensional Hilbert spaces to encode input features,
allowing more complex patterns to be represented with fewer parameters than
classical counterparts \cite{havlivcek2019supervised,schuld2019quantum}. This
is particularly beneficial when applied to non-stationary or noisy signals
frequently encountered in industrial maintenance and prognostic
applications \cite{emmanoulopoulos2022quantummachinelearningfinance,sagingalieva2025photovoltaic}.

Hybrid quantum--classical neural networks (HQNNs) combine classical and
quantum machine learning within a single
framework \cite{kordzanganeh2023exponentially,arthur2022hybridquantumclassicalneuralnetwork,haboury2024information,Bischof2025Hybrid,SUN2025130226,patapovich2025superposed}.
The quantum component plays a specific role in feature encoding or
transformation while leaving backpropagation and parameter optimisation to
classical routines \cite{broughton2021tensorflowquantumsoftwareframework}.
Preliminary studies indicate that HQNNs can match or exceed the performance
of their classical counterparts \cite{haboury2023supervised,sagingalieva2025hybrid},
often exhibiting greater resilience to overfitting \cite{abbas2021power,Berberich2024training}.
The inherent parameter efficiency of HQNNs further enhances their suitability
for scenarios involving complex, high-dimensional inputs or limited data
availability, conditions that are typical in predictive maintenance.

Data-driven prognostics of aero-engines has matured over more than a decade,
with the NASA C-MAPSS dataset \cite{cmapss,saxena2008turbofan} serving as the
de facto benchmark. Early studies applied conventional machine-learning
models (Random Forest, LASSO regression, Support Vector Machines,
$k$-nearest-neighbour regression, and Gradient
Boosting \cite{zhang2016multiobjective}), which provide interpretable baselines
but cannot natively model long-range temporal structure in the sensor signals.
Deep-learning approaches, including Multilayer Perceptrons, Convolutional Neural
Networks \cite{sateesh2016deep}, and LSTM-based recurrent
models \cite{zheng2017long}, explicitly capture temporal dependencies yet still
saturate at moderate RMSE on the relatively small FD001 subset. The current
state of the art combines complementary architectures with sophisticated feature
preprocessing: Temporal Convolutional Networks coupled with
Transformers~\cite{wang2021remaining}, attention-aware CNN--LSTM
models \cite{deng2024prediction}, broad-learning systems fused with
TCNs~\cite{yu2021prediction}, FCLCNN--LSTM hybrids \cite{peng2021remaining}, and
Auto-RUL-augmented LSTMs \cite{asif2022deep}, achieving the lowest reported RMSE
on FD001 at the cost of substantially more parameters and preprocessing~complexity.

While quantum--classical hybrids have been explored in
finance \cite{emmanoulopoulos2022quantummachinelearningfinance}, energy
forecasting \cite{sagingalieva2025photovoltaic,kurkin2025forecasting}, materials
science \cite{laskaris2026multi}, and
healthcare \cite{lusnig2024hybrid}, their use
in industrial prognostics is still at an early stage. A small but growing body of
work has begun to apply quantum, hybrid, and quantum-inspired methods to the
C-MAPSS benchmark, and QLSTM variants such as that of
Chen et al.~\cite{chen2022quantum} provide the algorithmic foundation. To the
best of our knowledge, however, gate-based hybrid recurrent models have not yet
been systematically benchmarked at matched parameter counts against the classical
and joint state of the art on standard C-MAPSS protocols. The present work
contributes to this emerging area with such a comparison and a circuit-level
analysis of the quantum layer.

This paper proposes and evaluates a Hybrid Quantum
Recurrent Neural Network (HQRNN) tailored for jet-engine RUL forecasting on
the NASA Commercial Modular Aero-Propulsion System Simulation (C-MAPSS)
benchmark \cite{saxena2008turbofan,cmapss}. Our specific contributions are as follows:

\begin{enumerate}
    \item We introduce a Quantum Long Short-Term Memory (QLSTM) architecture
    in which the linear transformation within each LSTM gate is replaced by a
    Quantum Depth-Infused (QDI)
    circuit \cite{sagingalieva2023hybrid,anoshin2024hybrid,lusnig2024hybrid,chen2022quantum,sagingalieva2025photovoltaic}.
    \item We empirically demonstrate that, at matched parameter budgets, the
    HQRNN improves average RMSE and MAE by approximately $5\%$ over classical
    LSTM-based RNNs on the C-MAPSS FD001 subset, despite using fewer
    trainable parameters in some configurations.
    \item We benchmark the HQRNN against classical machine-learning baselines
    (Random Forest, LASSO, SVM, KNR, Gradient Boosting), simple neural-network
    baselines (MLP, CNN, LSTM), and several recent joint deep-learning
    architectures, situating the proposed method within the current
    state-of-the-art landscape for the C-MAPSS~benchmark.
    \item We analyse the QDI quantum circuit through three complementary
    perspectives (ZX calculus, Fisher information, and Fourier
    expressivity) to verify that the chosen circuit is parameter-efficient,
    well-trainable, and capable of representing high-frequency components
    relevant to degradation modelling.
\end{enumerate}

The remainder of the paper is organised as follows.
Section~\ref{sec:dataset} describes the C-MAPSS dataset and the
prognostic problem formulation. Section~\ref{sec:model} introduces the HQRNN
architecture and the QDI circuit. Section~\ref{sec:training} details the
experimental setup and reports the empirical results.
Section~\ref{sec:qcirc_analysis} analyses the quantum circuit.
Section~\ref{sec:discussion} discusses implications for prognostics pipelines,
and Section~\ref{sec:conclusion} concludes the paper.

\section{Dataset and Problem Formulation}\label{sec:dataset}

The NASA C-MAPSS dataset \cite{cmapss} is a widely used multivariate
time-series benchmark for analysing gas-turbine engine degradation. In this
work we employ the FD001
 subset, which contains run-to-failure data
for a fleet of $100$ turbofan engines. Each engine progresses from a nominal
state to the point of failure, providing sensor measurements over multiple
cycles.

The C-MAPSS data are produced by a high-fidelity thermo-dynamical simulation of
a commercial turbofan engine \cite{cmapss,saxena2008turbofan}. Each trajectory
starts from a healthy engine with small, randomly assigned initial wear and
manufacturing variation and degrades under a prescribed fault until a failure
threshold is reached, with the recorded sensor outputs corrupted by measurement
noise. The FD001 subset corresponds to a single operating condition and a single
fault mode (high-pressure-compressor degradation), which makes it the most
controlled, and hence the most widely reported, of the four C-MAPSS subsets.

The dataset comprises four main feature groups:

\begin{enumerate}
    \item An engine identifier (ranging from $1$ to $100$);
    \item A time index in cycles;
    \item Three operational settings;
    \item Twenty-one sensor measurements.
\end{enumerate}

For simplicity, we exclude the operational settings and focus on the sensor
measurements for predicting RUL. More sophisticated feature-selection schemes,
including hybrid quantum--classical selectors~\cite{lopatkin2026quantum}, could
further prune the input set, but are beyond the scope of this study.

The C-MAPSS benchmark is distributed already partitioned into a training set and
a test set. This dataset-level split is the one provided by NASA and should not
be confused with the train/validation split that we later carve out of the
training set for model development (Section~\ref{sec:training}). In the provided
training set, each engine's sensor measurements are recorded until failure,
whereas in the test set the measurements end at an arbitrary cycle prior to
failure. The prognostic task is to predict the number of remaining operational
cycles before failure for each test engine, and a vector of ground-truth RUL
values for the test data is provided for evaluation.

We design our predictive model as follows. The model receives a fixed-size
window of consecutive engine cycles as input and yields a single estimated
RUL value as output. This means that for each entry in the training dataset
there should be a corresponding target RUL value.

\begin{figure}[t]
    \centering
    \includegraphics[width=0.85\linewidth]{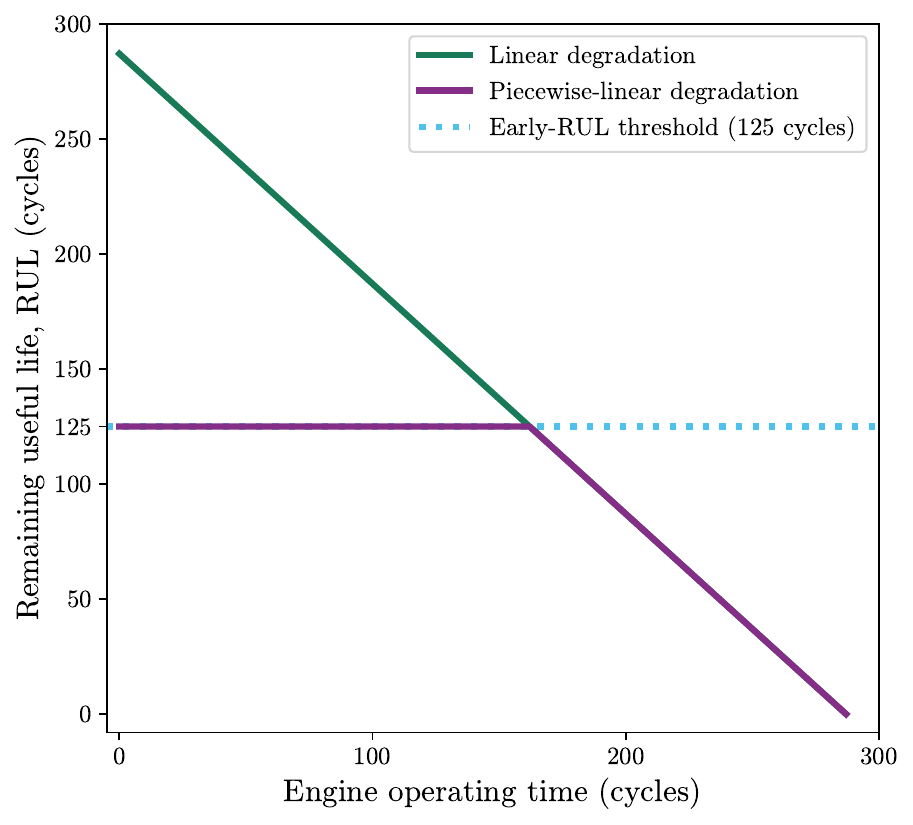}
    \caption{Remaining useful life (RUL) of engine~2 (FD001)
    under the linear and piecewise-linear degradation models. Both axes are
    measured in engine operating cycles but represent complementary quantities:
    the horizontal axis is the elapsed operating time~$t$ (the number of
    cycles the engine has already run), whereas the vertical axis is the
    remaining useful life until failure,
    $\mathrm{RUL}(t) = T_\mathrm{life} - t$ (here the engine fails at
    $T_\mathrm{life} = 287$ cycles). The dotted line marks the early-RUL
    threshold: this work adopts the piecewise-linear model, which caps the RUL
    target at $125$ cycles, so that the network is only asked to track the RUL
    once the engine approaches end of life.}
    \label{fig:degradation_model}
\end{figure}

Creation of these RUL values depends on the degradation model. A linear
degradation model decreases the RUL linearly to zero as the engine approaches
failure. In this work we instead use a piecewise-linear degradation
model. It assumes that the engine operates normally for an initial period,
during which the RUL is held constant (the ``early-RUL'' regime), and then
develops a fault, after which its useful life decreases linearly
(see Figure~\ref{fig:degradation_model}). We set the early-RUL threshold to
$125$ cycles, based on the minimum run length of $128$~cycles observed in
the training data.

Upon inspecting the training data, we found that seven of the twenty-one sensor
channels ($s_1, s_5, s_6, s_{10}, s_{16}, s_{18}, s_{19}$) remain constant across
all cycles and therefore carry no predictive information. Together with the three
operational-setting channels, these are removed, leaving $14$ informative sensor
channels ($s_2, s_3, s_4, s_7, s_8, s_9, s_{11}, s_{12}, s_{13}, s_{14}, s_{15},
s_{17}, s_{20}, s_{21}$). Each retained channel is standardised to zero mean and
unit variance ($z$-score normalisation); the scaler statistics are estimated on
the training data only and then applied unchanged to the test data, so no test
information influences preprocessing.

Because our proposed architecture relies on LSTM-like layers, we segment
each time series into windows of length $30$ cycles. Each window is therefore
a $30 \times 14$ matrix of sensor values. The network takes this window as
input and produces a single RUL estimate for the cycle immediately following
the window, yielding a supervised learning setup for the engine-degradation
problem.

\section{Hybrid Quantum Recurrent Neural Network}\label{sec:model}

The proposed HQRNN comprises a stack of Quantum Long Short-Term Memory
(QLSTM) layers followed by classical fully connected layers for final
regression. The QLSTM layer is derived from the conventional LSTM by
substituting the linear transformations in each of the four gates (forget,
input, update, output) with a Quantum Depth-Infused (QDI)~layer.

Classical deep neural networks often bias their learning towards
lower-frequency components, a phenomenon sometimes referred to as the
``F-Principle'' \cite{xu2019training}. Angle-encoded quantum circuits such as the
QDI layer can be interpreted as operating in a Fourier-transformed feature
space \cite{schuld2021effect}, which may make it easier to represent
higher-frequency components. We treat this as a motivating inductive bias rather
than a proven advantage; the empirical comparisons in
Section~\ref{sec:training} and the circuit-level analysis in
Section~\ref{sec:qcirc_analysis} are what we rely on.

\subsection{Architecture Overview}\label{sec:arch}

\begin{figure*}[t]
    \centering
    \includegraphics[width=\linewidth]{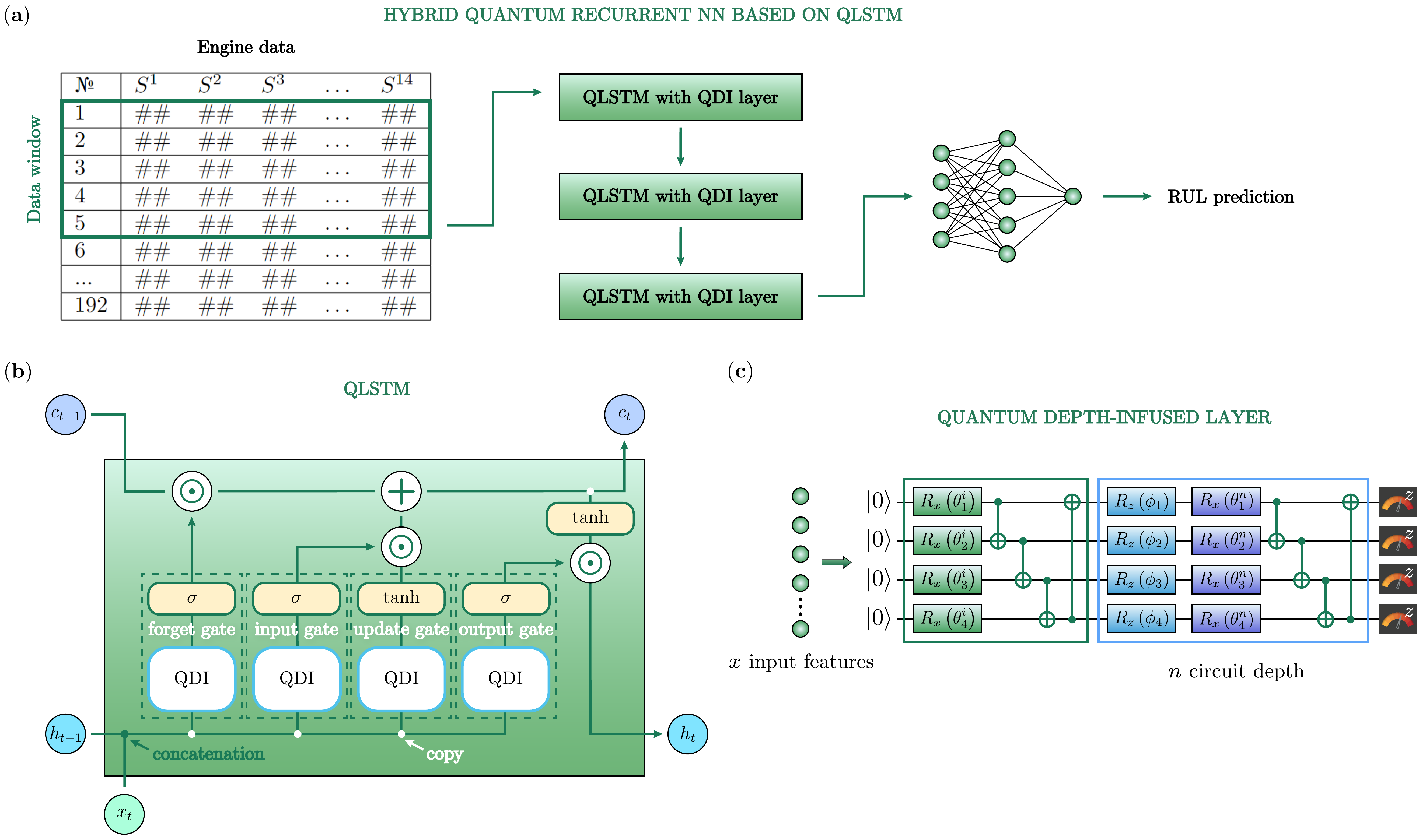}
    \caption{(\textbf{a}) HQRNN 
 model pipeline. A data window of size $W \times 14$
    is processed through three stacked QLSTM layers followed by classical
    dense layers, yielding a single RUL value. The dimensions of the QLSTM
    layers are $32$, $16$, and $8$, while the dense layers transition from
    $8 \times W$ to $16$, $16$ to $32$, and finally $32$ to $1$. The \#\#
    symbols in the engine-data table denote numeric sensor readings.
    (\textbf{b}) Structure of the QLSTM layer. Conventional linear transformations
    are replaced by QDI layers for each of the four LSTM gates (forget,
    input, update, output).
    (\textbf{c}) Schematic of the QDI layer used in the QLSTM. Each input feature is
    encoded via $R_z$ gates parameterised by that feature on a four-qubit quantum
    circuit. The variation part consists of trainable $R_x$ rotations and CNOT
    gates. The blue block repeats $n = 1$ time. The observable is the Pauli
    $Y$ matrix.}
    \label{fig:diagram}
\end{figure*}

Figure~\ref{fig:diagram} provides an overview of the HQRNN model and its
constituent layers. Figure~\ref{fig:diagram}a shows the complete pipeline:
an input window of sensor measurements of size $W \times 14$ (where $W = 30$
is the window length and $14$ is the number of sensor features) is passed
through three stacked QLSTM layers with latent dimensions of $32$, $16$, and
$8$ respectively. The output of the QLSTM stack is then fed into classical
dense layers that reduce the dimension from $8 \times W$ to $16$, then $32$,
and finally to a single scalar value representing the predicted RUL.
Figure~\ref{fig:diagram}b depicts the internal structure of a single QLSTM
layer, highlighting the replacement of each conventional linear gate
transformation with an individual QDI layer.

Crucially, each gate of the QLSTM is not a bare quantum circuit but a
classical--quantum--classical composition, and the QDI circuits never receive
raw sensor readings. Let 
 $x_t \in \mathbb{R}^{d_\mathrm{in}}$ denote the input at
time step $t$ of a QLSTM layer with hidden size $d_h$, and let
$h_{t-1}, c_{t-1} \in \mathbb{R}^{d_h}$ be the previous hidden and cell states.
The cell first forms a classical linear input projection
\begin{equation}
    v_t = W_x\, x_t + W_h\, h_{t-1}, \qquad v_t \in \mathbb{R}^{4m},
    \label{eq:preact}
\end{equation}
where $m = n_q\, L = 4$ is the QDI input width ($n_q = 4$ qubits and $L = 1$
encoding repetition), and $W_x \in \mathbb{R}^{4m \times d_\mathrm{in}}$,
$W_h \in \mathbb{R}^{4m \times d_h}$ are trainable weight matrices that together
form the input projector. The vector
$v_t$ is split into four gate input projections
$a_t^{f}, a_t^{i}, a_t^{g}, a_t^{o} \in \mathbb{R}^{m}$, one per gate. Each gate
passes its input projection through a QDI circuit and a shared linear
readout $\mathrm{clayer\_out}: \mathbb{R}^{n_q} \to \mathbb{R}^{d_h}$,
\begin{equation}
    z_t^{\gamma} = \mathrm{clayer\_out}\!\left(\mathrm{QDI}_{\gamma}(a_t^{\gamma})\right),
    \qquad \gamma \in \{f, i, g, o\},
    \label{eq:gate}
\end{equation}
where $\mathrm{QDI}_{\gamma}: \mathbb{R}^{m} \to \mathbb{R}^{n_q}$ is the
four-qubit circuit of Section~\ref{sec:qdi}. The standard LSTM recurrence then
follows,
\begin{align}
    f_t &= \sigma(z_t^{f}), \quad i_t = \sigma(z_t^{i}), \quad
           g_t = \tanh(z_t^{g}), \quad o_t = \sigma(z_t^{o}), \\
    c_t &= f_t \odot c_{t-1} + i_t \odot g_t, \qquad
    h_t  = o_t \odot \tanh(c_t),
\end{align}
with $\sigma$ the logistic sigmoid and $\odot$ the elementwise product.
Algorithm~\ref{alg:qlstm} summarises one cell step, and
Table~\ref{tab:shapes} lists the tensor shapes for a batch of $B$ windows.
Consequently, the four values encoded into each QDI circuit are classical linear
projections of the concatenated input and hidden state.

\begin{algorithm}[H]
\caption{QLSTM cell (one time step of one layer).}
\label{alg:qlstm}
\begin{algorithmic}[1]
\Require input $x_t \in \mathbb{R}^{d_\mathrm{in}}$; previous states
         $h_{t-1}, c_{t-1} \in \mathbb{R}^{d_h}$
\Statex \textbf{Parameters:} $W_x \in \mathbb{R}^{4m \times d_\mathrm{in}}$,
        $W_h \in \mathbb{R}^{4m \times d_h}$, shared readout
        $\mathrm{clayer\_out}: \mathbb{R}^{n_q} \!\to\! \mathbb{R}^{d_h}$, four
        circuits $\{\mathrm{QDI}_{\gamma}\}$ ($m = n_q L = 4$)
\State $v_t \gets W_x x_t + W_h h_{t-1}$
       \Comment{$v_t \in \mathbb{R}^{4m}$}
\State $(a_t^{f}, a_t^{i}, a_t^{g}, a_t^{o}) \gets \mathrm{split}(v_t)$
       \Comment{each $a_t^{\gamma} \in \mathbb{R}^{m}$}
\ForAll{$\gamma \in \{f, i, g, o\}$}
  \State $z_t^{\gamma} \gets \mathrm{clayer\_out}\!\left(\mathrm{QDI}_{\gamma}(a_t^{\gamma})\right)$
         \Comment{$\mathrm{QDI}_{\gamma}: \mathbb{R}^{m} \!\to\! \mathbb{R}^{n_q}$}
\EndFor
\State $f_t \gets \sigma(z_t^{f})$;\ \ $i_t \gets \sigma(z_t^{i})$;\ \
       $g_t \gets \tanh(z_t^{g})$;\ \ $o_t \gets \sigma(z_t^{o})$
\State $c_t \gets f_t \odot c_{t-1} + i_t \odot g_t$
\State $h_t \gets o_t \odot \tanh(c_t)$
\Ensure $h_t,\ c_t$
\end{algorithmic}
\end{algorithm}

\begin{table*}[t]
\caption{Tensor shapes within one QLSTM cell for a batch of $B$ windows, with
QDI input width $m = n_q L = 4$. In the reported model $n_q = 4$, $L = 1$, and
$(d_h^{(1)}, d_h^{(2)}, d_h^{(3)}) = (32, 16, 8)$ across the three stacked
layers.}
\label{tab:shapes}
\centering
\renewcommand{\arraystretch}{1.2}
\begin{tabular}{lll}
\toprule
\textbf{{Symbol}} & \textbf{{Definition}} & \textbf{{Shape}} \\
\midrule
$x_t$ & layer input at step $t$ & $B \times d_\mathrm{in}$ \\
$h_{t-1}$ & previous hidden state & $B \times d_h$ \\
$v_t = W_x x_t + W_h h_{t-1}$ & input projection & $B \times 4m\ (=\!B\times16)$ \\
$a_t^{\gamma}$ & gate input projection, $\gamma\!\in\!\{f,i,g,o\}$ & $B \times m\ (=\!B\times4)$ \\
$\mathrm{QDI}_{\gamma}(a_t^{\gamma})$ & four-qubit circuit output & $B \times n_q\ (=\!B\times4)$ \\
$z_t^{\gamma} = \mathrm{clayer\_out}(\cdot)$ & linear readout & $B \times d_h$ \\
$c_t,\ h_t$ & cell / hidden state & $B \times d_h$ \\
\bottomrule
\end{tabular}
\end{table*}

Because a cell evaluates four QDI circuits per time step, processing one
$W$-cycle window through the three stacked QLSTM layers requires
$4 \times W \times 3 = 360$ circuit evaluations for $W = 30$ (each circuit
evaluated once per gate, per time step, per layer). Every QDI circuit carries
exactly $8$ trainable rotation angles (Section~\ref{sec:qdi}), so the quantum
part of the model contains $4\ \text{gates} \times 3\ \text{layers} \times 8 =
96$ trainable angles--about $1.4\%$ of the $6793$ trainable parameters of the
reported HQRNN. The overwhelming majority of parameters therefore reside in the
classical $W_x$, $W_h$, shared readout, and dense-head weights; the quantum
circuits act as compact, fixed-width nonlinear gate transformations.
This organisation is not a ``made-up'' construct. The input projection
$v_t = W_x x_t + W_h h_{t-1}$ followed by a split into four gate input
projections is exactly the gate structure of the standard PyTorch (v2.13.0+cpu) 
 \texttt{LSTM}
cell that we use as the classical baseline; the only change introduced by the
HQRNN is that each gate's transformation is routed through a QDI circuit instead
of a bare linear map. The HQRNN is therefore a faithful, minimally modified
quantum extension of the reference LSTM architecture.

\subsection{Quantum Depth-Infused Circuit}\label{sec:qdi}

Figure~\ref{fig:diagram}c shows the design of the QDI layer in detail.
Before encoding the input data, a parameterised $R_x$ rotation is applied to
the initial state of the four qubits using the trained parameters as
rotation angles. These are combined with a ring of controlled NOT (CNOT)
gates (the orange operation block):
\begin{equation}
\begin{aligned}
R_x(\theta) &= \begin{pmatrix}
  \cos(\theta/2) & -i \sin(\theta/2)\\[2pt]
  -i \sin(\theta/2) & \cos(\theta/2)
\end{pmatrix},\\[4pt]
\mathrm{CNOT} &= \begin{pmatrix}
  1 & 0 & 0 & 0 \\
  0 & 1 & 0 & 0 \\
  0 & 0 & 0 & 1 \\
  0 & 0 & 1 & 0 \\
\end{pmatrix}.
\end{aligned}
\end{equation}

This is followed by an encoding block of $R_z$ rotations parameterised by the
input features (blue rectangles):
\begin{equation}
R_z(\phi) = \begin{pmatrix}
  e^{-i\phi/2} & 0\\
  0 & e^{i\phi/2}
\end{pmatrix}.
\end{equation}

The variation block of $R_x$ rotations combined with CNOT gates is then
repeated $n = 1$ time. The circuit is completed by measurement in the basis
of the Pauli-$Y$ eigenvectors.

With $n_q = 4$ qubits, depth $1$, and a single encoding repetition, the two
$R_x$ variational blocks, one before and one after the encoding layer, provide
$4 + 4 = 8$ trainable rotation angles, which are the only quantum parameters of
the QDI layer. The $R_z$ encoding gates are parameterised by the (classical) gate
input projection $a_t^{\gamma}$ of Equation~\eqref{eq:preact} and carry no trainable
weights, and the CNOT gates are fixed. The input to each circuit is thus a
four-dimensional classical projection rather than a raw sensor reading, and the
Pauli-$Y$ expectation values on the four wires form the circuit output.

For comparison, we also train a purely classical LSTM-based RNN that retains
the same overall structure but replaces each QDI layer with a standard
linear transformation.

\section{Experimental Setup and Results}\label{sec:training}

In our experiments the HQRNN takes a $30 \times 14$ window of sensor data
and outputs the corresponding RUL. We split the training dataset using an
$80/20$ train/validation partition, holding out $20\%$ for validation, while
the test dataset is used exclusively for final performance assessment.

The training objective is to minimise the mean squared error (MSE) loss:
\begin{equation}
    \mathrm{MSE} = \frac{1}{N} \sum_{i=1}^{N} \bigl(\hat{y}_i - y_i\bigr)^2,
\end{equation}
where $\hat{y}_i$ and $y_i$ are the predicted and ground-truth RUL values
for the $i$-th sample. In addition to MSE, we evaluate root-mean-squared
error (RMSE) and mean absolute error (MAE) on the validation set to gain
further insight into each model's predictive performance.

Using the Adam optimiser with a batch size of $128$ and a learning rate of
$0.001$, we train both the HQRNN and a purely classical RNN for $20$ epochs.
The two models are configured with an identical number of parameters in
matched pairs to ensure a fair comparison.

The quantum layers are implemented in PennyLane and executed on the
\texttt{lightning.qubit} state-vector simulator; circuit gradients are obtained
by the adjoint differentiation method and integrated into the PyTorch autograd
graph through the \texttt{torch} interface. Both models use a global batch size
of $128$; the quantum models are trained with $16$-way \texttt{gloo} data
parallelism (a per-process batch of $8$), while the classical baselines are
trained single-process. Because the QDI layers are parameter-compact, an HQRNN
and a classical RNN that share the same layer widths do not share the same
parameter count: at the widths used here, the HQRNN has $6793$ trainable
parameters, whereas the identically shaped classical baseline
RNN-$32$-$16$-$8$-$16$-$32$ has $14{,}609$. We therefore compare against two
families of classical baselines: (i) an identical-hyperparameter RNN that
reuses the HQRNN's layer widths; (ii) a matched-parameter RNN whose
widths are reduced (RNN-$20$-$16$-$4$-$8$-$16$) so that its $6793$ trainable
parameters equal the HQRNN's (see Table~\ref{tab:results}). Comparison~(i)
controls for the shape of the architecture and comparison~(ii) for its capacity;
in particular, holding the parameter count equal in case~(ii) ensures that any
performance difference between a matched pair is not attributable to one model
simply being larger.

To rule out information leakage, we note that the reported RMSE and MAE are
computed exclusively on the held-out C-MAPSS test engines
(\texttt{test\_FD001} with the provided \texttt{RUL\_FD001} labels), for which
the last observation window of each engine is used. The $80/20$ partition
described above splits only the training data into train and validation subsets
and never involves the test engines. Because training runs for a fixed $20$
epochs without early stopping and the final-epoch checkpoint is the one
evaluated, there is no selection path from the validation set to the reported
test metric. The identical preprocessing and data-handling pipeline is applied to
the quantum and classical models.

Each model is trained with $10$ different random seeds, and we report the
mean of the resulting per-run test metrics. As shown in
Table~\ref{tab:results}, the HQRNN outperforms the classical RNN in both
RMSE and MAE across all tested parameter configurations. Although the
classical model achieves slightly lower losses on the training and
validation sets, the HQRNN demonstrates superior generalisation on the test
set. This observation is consistent with theoretical and empirical results
suggesting that quantum models may generalise better in data-scarce
scenarios \cite{caro2022generalization,schuld2021effect}: in our setup, the
training fleet comprises only $100$~engines.

A possible explanation lies in the quantum circuit's capacity to capture
higher-frequency components of the underlying
function \cite{xu2019training}, which becomes beneficial when the available
context (the fleet of engines) is small. In practical terms, our findings
indicate that the HQRNN may provide robust RUL predictions even when only
short segments of sensor readings are accessible, a particularly valuable
property for real-world aerospace applications where complete sensor
histories are often unavailable or expensive to obtain.

\begin{table*}[t]
\caption{Comparison of RNN and HQRNN models on the C-MAPSS
FD001 test set. Mean RMSE/MAE are reported as mean $\pm$ standard deviation over
$10$ seeds (s125--s134); the ``Best RMSE''/``Best MAE'' columns list the strongest
single run. Bold values denote the best mean (RMSE, MAE) and the best single run.}
\label{tab:results}
\centering
\begin{tabular}{lccccc}
\toprule
\textbf{{Model}}              & \textbf{{Mean RMSE $\pm$ std}}        & \textbf{{Best RMSE}}        & \textbf{{Mean MAE $\pm$ std}}         & \textbf{{Best MAE}}         & \textbf{{No.\ of Params 
}} \\
\midrule
HQRNN              & $\mathbf{15.46 \pm 0.89}$  & $\mathbf{14.78}$          & $\mathbf{12.25 \pm 0.84}$  & $\mathbf{11.51}$          & $6793$          \\
RNN-32-16-8-16-32  & $16.71 \pm 0.64$           & $15.68$          & $13.18 \pm 0.55$           & $12.19$ & $14{,}609$         \\
RNN-20-16-4-8-16   & $16.37 \pm 0.46$           & $15.73$          & $12.89 \pm 0.23$           & $12.51$          & $6793$          \\
RNN-16-8-4-8-16    & $16.56 \pm 0.63$           & $15.52$          & $13.03 \pm 0.42$           & $12.36$          & $4233$          \\
RNN-8-4-2-4-8      & $29.72 \pm 20.81$          & $15.07$ & $24.52 \pm 17.98$          & $12.20$          & $\mathbf{1349}$ \\
\bottomrule
\end{tabular}
\end{table*}

To characterise run-to-run variability we report the mean $\pm$ standard
deviation of the test metrics over the ten seeds in Table~\ref{tab:results}.
Because the quantum and classical models are trained from the same ten
seeds, we compare matched pairs with a paired (by-seed) two-sided $t$-test, which
removes seed-to-seed variance and is more powerful than comparing marginal
intervals. Against the matched-parameter baseline RNN-20-16-4-8-16 ($6793$~parameters), the HQRNN lowers the mean test RMSE from $16.37$ to $15.46$
($5.6\%$) and the mean MAE from $12.89$ to $12.25$ ($5.0\%$); the RMSE
improvement is statistically significant ($t(9) = 2.80$, $p = 0.021$; Wilcoxon
signed-rank $p = 0.027$), and the HQRNN attains the lower RMSE on nine of the ten
seeds. The corresponding $95\%$ confidence intervals are $[14.82, 16.10]$ (HQRNN)
and $[16.04, 16.69]$ (RNN-20-16-4-8-16); although these marginal intervals
overlap slightly, the paired test is significant because the two models respond
consistently across seeds. The HQRNN is likewise significantly better than the
larger RNN-32-16-8-16-32 ($t(9) = 3.30$, $p = 0.009$) and the smaller
RNN-16-8-4-8-16 ($t(9) = 3.67$, $p = 0.005$). We exclude the smallest baseline,
RNN-8-4-2-4-8 ($1349$ parameters), from this significance comparison: three of
its ten runs failed to converge (test RMSE $> 50$), inflating its mean and
standard deviation ($29.72 \pm 20.81$) and indicating that a model this small
trains unstably rather than competitively.

\begin{figure*}[t]
    \centering
    \includegraphics[width=1.8\columnwidth]{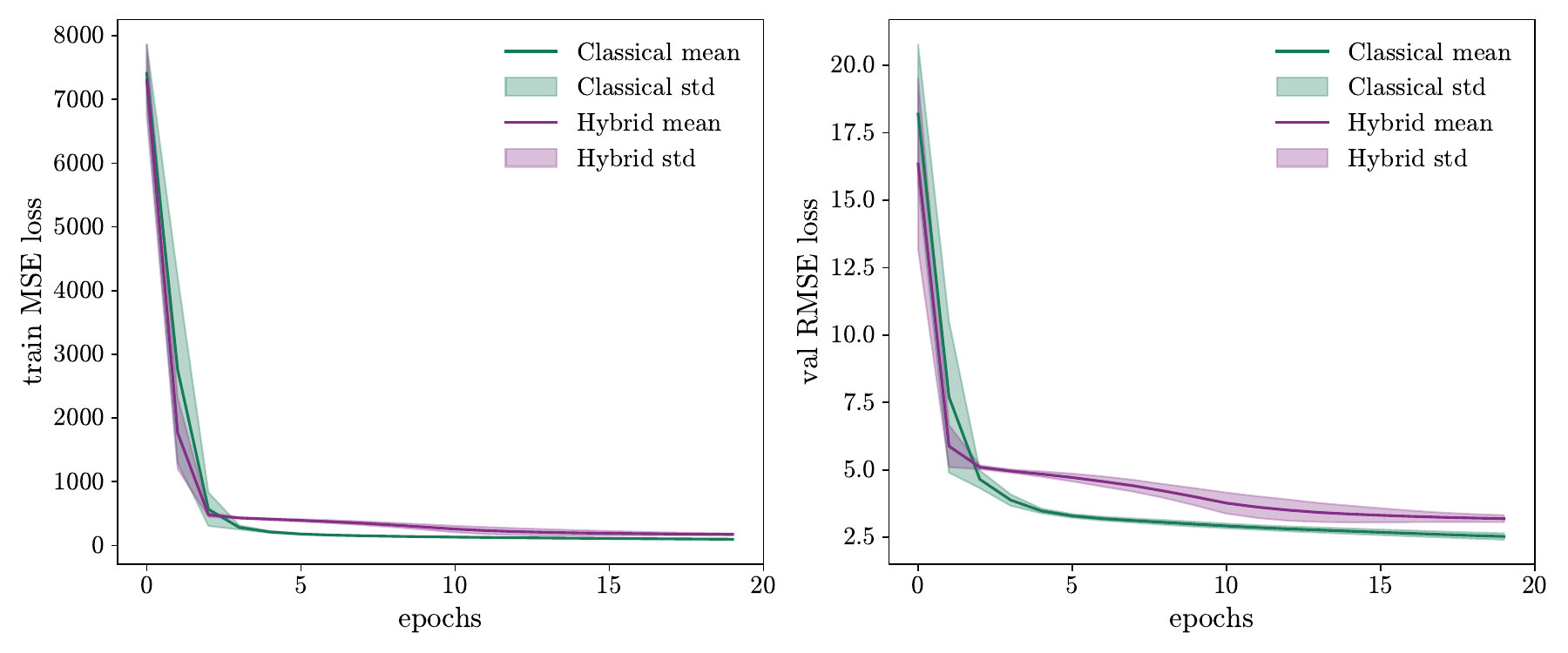}
    \caption{Convergence of the classical RNN (``Classical'') and
    the HQRNN (``Hybrid'') over the twenty training epochs: training MSE loss
    (left) and validation RMSE (right), shown as the mean (solid line) $\pm$ one
    standard deviation (shaded band) across the ten seeds. Both curves plateau
    well before epoch~20, indicating convergence under the fixed-epoch schedule.}
    \label{fig:train_curves}
\end{figure*}

Figure~\ref{fig:train_curves} shows the training MSE loss and the validation
RMSE over the twenty epochs, as mean $\pm$ standard deviation across the ten
seeds, for the HQRNN and its matched-parameter classical RNN
(RNN-20-16-4-8-16). Both models converge rapidly: the training loss and the
validation RMSE drop sharply within the first few epochs and are essentially
flat well before epoch~20, confirming that the fixed twenty-epoch schedule is
sufficient on FD001 and that neither model is undertrained. Consistent with
Table~\ref{tab:results}, the classical RNN reaches a marginally lower
validation RMSE while the HQRNN generalises better on the held-out
test engines; note that the validation RMSE is evaluated on overlapping
windows and is therefore optimistic relative to the engine-level test metric.

\label{sec:cost}%

We profiled the training and inference cost of every model on CPU, using the
same PennyLane \texttt{lightning.qubit} state-vector backend and adjoint
gradients as in training (i.e.,\ noiseless simulation). Per-epoch training times
are obtained by timing individual $128$-sample batches and scaling by the $111$
batches of a full FD001 epoch; every figure is a mean over five repetitions. Table~\ref{tab:cost} summarises the results.

\begin{table*}[t]
\caption{Computational cost on CPU (PennyLane
\texttt{lightning.qubit}, noiseless state-vector simulation). Training and
inference figures are mean $\pm$ standard deviation over five repetitions
(warm-up excluded); the classical models' timing variability rounds below the
displayed precision. Per-epoch training times are extrapolated from the measured
per-batch times ($128$-sample batches, $111$ batches per FD001 epoch); the
full-schedule ($20$-epoch) times quoted in the text follow by multiplication.
``Infer'' is the mean inference time per test window and ``Peak mem'' the maximum
resident set size. Each HQRNN forward pass evaluates $360$ four-qubit circuits
per window ($\approx\!5.1\times10^{6}$ per training epoch); the classical models
use none.}
\label{tab:cost}
\centering
\begin{tabular}{lcccc}
\toprule
\textbf{{Model}} & \textbf{{Params}} & \textbf{{Train (s/epoch)}} & \textbf{{Infer (ms/win)}} & \textbf{{Peak Mem (GB)}} \\
\midrule
HQRNN (\texttt{Z\_basic\_Y}) & $6793$  & $4189 \pm 22$   & $242 \pm 2$      & $2.3$  \\
RNN-32-16-8-16-32            & $14609$ & $2.01 \pm 0.06$ & $0.047 \pm 0.000$ & $0.49$ \\
RNN-20-16-4-8-16             & $6793$  & $2.00 \pm 0.06$ & $0.049 \pm 0.000$ & $0.49$ \\
RNN-16-8-4-8-16              & $4233$  & $1.50 \pm 0.01$ & $0.050 \pm 0.000$ & $0.49$ \\
RNN-8-4-2-4-8                & $1349$  & $1.35 \pm 0.04$ & $0.048 \pm 0.000$ & $0.50$ \\
\bottomrule
\end{tabular}
\end{table*}

Simulating the quantum circuits dominates the cost. Each HQRNN forward pass
evaluates $360$ four-qubit circuits (Section~\ref{sec:arch}), i.e.,\ about
$5.1\times10^{6}$ logical circuit evaluations per epoch over the $14{,}184$
training windows. These evaluations are not run one at a time: the $128$-sample
batch is processed together through PennyLane's parameter broadcasting and the
PyTorch interface, so the timings reported here already reflect this batched
execution and are not a naive product of the evaluation count and a per-circuit
cost. Even so, a single-process training epoch takes
$\approx\!4.2\times10^{3}$~s, against $\approx\!2$~s for the matched-parameter
classical RNN: a factor of roughly $2\times10^{3}$. Inference costs $\approx\!0.24$~s per engine window
versus $\approx\!5\times10^{-5}$~s classically, and peak resident memory is
$\approx\!2.3$~GB versus $\approx\!0.5$~GB. This overhead comes from
classically simulating
 the quantum circuits, not from any intrinsic
property of quantum computation. The cost of such simulation grows exponentially
with the number of qubits, which is one reason we keep the circuits to four
qubits; on real quantum hardware, the per-circuit cost would scale very
differently.

Two practical consequences follow. First, the reported experiments used $16$-way
\texttt{gloo} data parallelism, which reduces the wall-clock time of a full
$20$-epoch training run to an estimated $\approx\!1.5$~h, against
$\approx\!23$~h single-process (both obtained from the per-epoch time of
Table~\ref{tab:cost}). Second, because the circuits use only four qubits, they
can be simulated exactly on an ordinary classical computer; the HQRNN can
therefore be trained and deployed today without access to a quantum processor.

We therefore state the scope of our claims precisely. The advantage demonstrated
in this work is one of parameter efficiency and predictive accuracy: at a
matched or smaller weight budget, the HQRNN attains a lower test error than
the classical RNN (Section~\ref{sec:training}). It is not an advantage in
computational efficiency. Under present-day classical simulation the HQRNN is
about $2\times10^{3}$ times slower to train, about $5\times10^{3}$ times slower
per inference, and roughly $5\times$ more memory-intensive than the
matched-parameter classical RNN. On modern simulators, therefore, the HQRNN
provides no advantage in speed or resource use: the model is more compact in its number of weights and more accurate,
but markedly more expensive to run on simulators. Its practical case rests on that compactness in
data-scarce regimes and on the prospect of native quantum execution once the
hardware matures (Section~\ref{sec:nisq}).

\section{Quantum Circuit Analysis}\label{sec:qcirc_analysis}

This section examines the QDI layer used in the QLSTM network through three
complementary perspectives: redundancy analysis with ZX calculus,
trainability analysis using Fisher information, and expressivity analysis
via Fourier-series decomposition.

With one exception, these analyses probe the isolated QDI circuit under
random or Gaussian inputs rather than the fully trained HQRNN; the exception is
the Fisher-information analysis, which we additionally repeat on the empirical
input distribution recorded from the trained models
(Section~\ref{sec:fisher}). They should be read as necessary, but not
sufficient, indicators that the circuit is compact, optimisable, and expressive,
and they do not, on their own, establish that the empirical improvement in
Section~\ref{sec:training} originates from a quantum-mechanical effect.

\subsection{Redundancy Analysis: ZX Calculus}\label{sec:zx}

ZX calculus is a graphical language for representing and simplifying quantum
circuits \cite{coecke2011interacting,van2020zx}. It employs ``spider'' nodes,
enabling the analysis and optimisation of quantum gates through well-defined
algebraic rewriting rules. By using ZX techniques, one can in principle
reduce the number of parameters and gates in a quantum circuit without
changing its overall functionality \cite{peham2022equivalence,wang2024differentiating}.

\begin{figure*}[ht!]
    \centering
    \includegraphics[width=1.8\columnwidth]{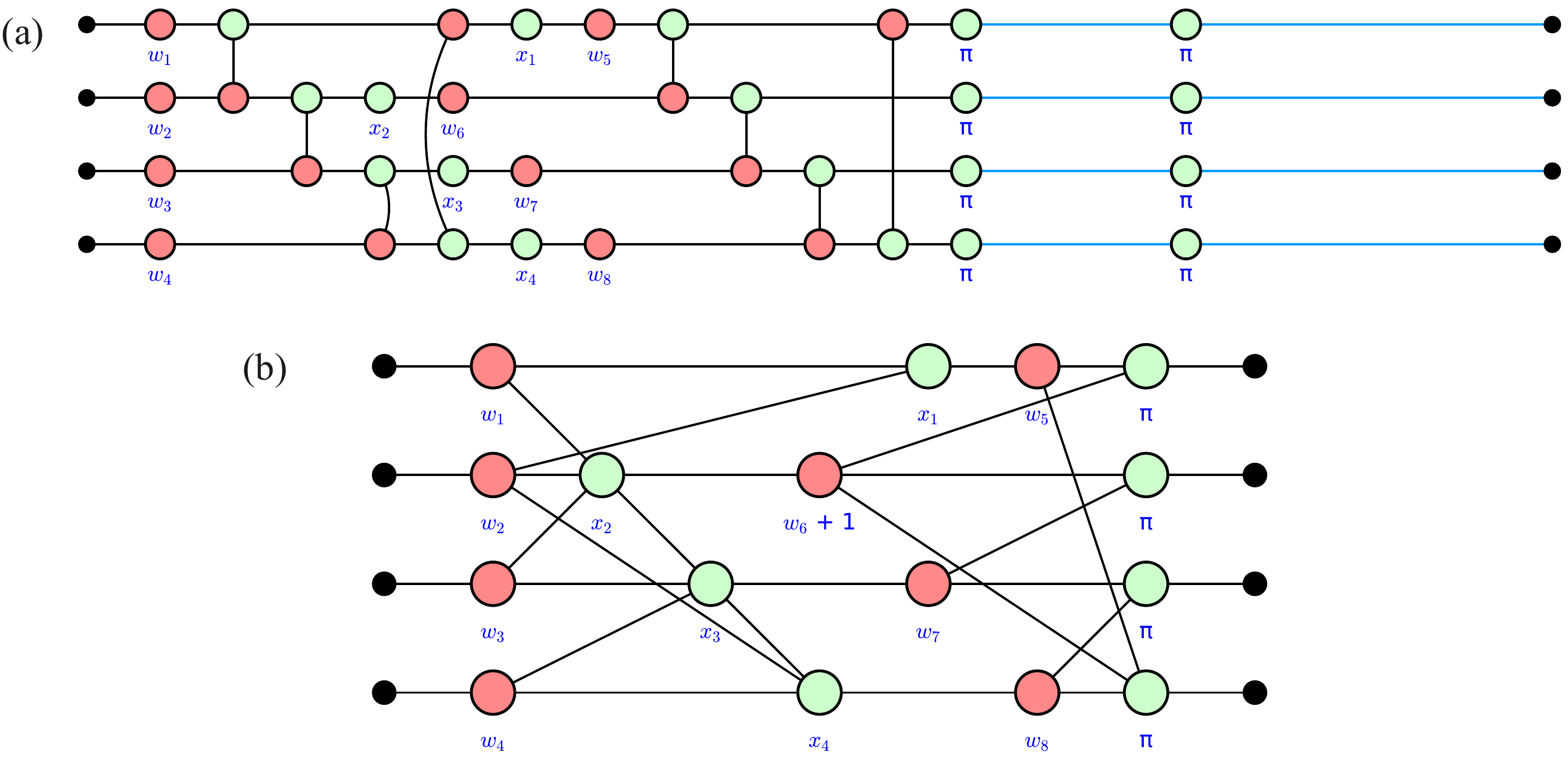}
    \caption{(\textbf{{a}}) A QDI layer before applying ZX-based parameter reduction.
    (\textbf{{b}}) The ZX-reduced QDI layer structure with rearranged weights. Despite
    these simplifications, no parameters can be removed without affecting
    the layer's functionality.}
    \label{fig:ZX}
\end{figure*}

Figure~\ref{fig:ZX} illustrates the original QDI circuit (a) and its
ZX-optimised form (b). The primary modifications involve rearranging certain
weights and merging CNOT gates with them. Under the rewrite rules we applied, all
$8$ parameters of the original circuit remain essential and no further parameter
reduction was found without altering the circuit's behaviour. The QDI structure
thus appears ZX-irreducible under these rules, consistent with efficient
parameter usage.

\subsection{Trainability: Fisher Information}\label{sec:fisher}

In supervised machine learning, a model $h_\theta(\hat{x})$ is trained on a
labelled dataset $\mathcal{D} = \{(x_i, y_i)\}_{i=1}^{N}$, where the model
parameters $\theta = (\theta^1, \theta^2, \ldots, \theta^n)$ define a
conditional probability distribution $p(y \mid x, \theta)$ that can also be
expressed via the joint distribution $p(x, y \mid \theta)$ as
$p(y \mid x, \theta) = p(x, y \mid \theta) / p(x)$. The set of all such
distributions for different $\theta$ forms a manifold
$\mathcal{M} = \{p(y \mid x, \theta), \,\theta \in \Theta\}$.

Defining the score function as
$s(\theta) = \nabla_\theta \log p(y \mid x, \theta)$, the Fisher Information
Matrix (FIM) is the covariance of the score \cite{amari1998gradient}:
\begin{equation}
    F(\theta)= \mathrm{Cov}\bigl[s(\theta)\bigr] =
    \mathbb{E}_{p(x,y\mid\theta)}\bigl[s(\theta)\,s(\theta)^{\!\top}\bigr].
    \label{eq:fisher}
\end{equation}
In practice the FIM is approximated by the sample average. Moreover, the
``volume'' of the manifold $\mathcal{M}$ can be evaluated as
$V \propto \int_{\Theta} \!\sqrt{\det F(\theta)}\, d\theta$, whose logarithm
corresponds to the effective dimension introduced
in \cite{berezniuk2020scale} and serves as a measure of model complexity.

A significant motivation for analysing the FIM is its connection to the
barren-plateau phenomenon, in which training gradients vanish exponentially with
system size and thereby obstruct optimisation \cite{mcclean2018barren}.
Barren plateaus are therefore detrimental to learning, and a trainable circuit
is one that avoids them. In terms of the FIM, a large fraction of
near-zero eigenvalues is the signature of a barren plateau and hence of poor
trainability, whereas an eigenvalue spectrum spread away from zero indicates
informative gradients and favourable trainability \cite{abbas2021power}. A
well-conditioned circuit is thus expected to show a negligible fraction of
near-zero eigenvalues.

\begin{figure*}[ht!]
    \centering
    \includegraphics[width=1.8\columnwidth]{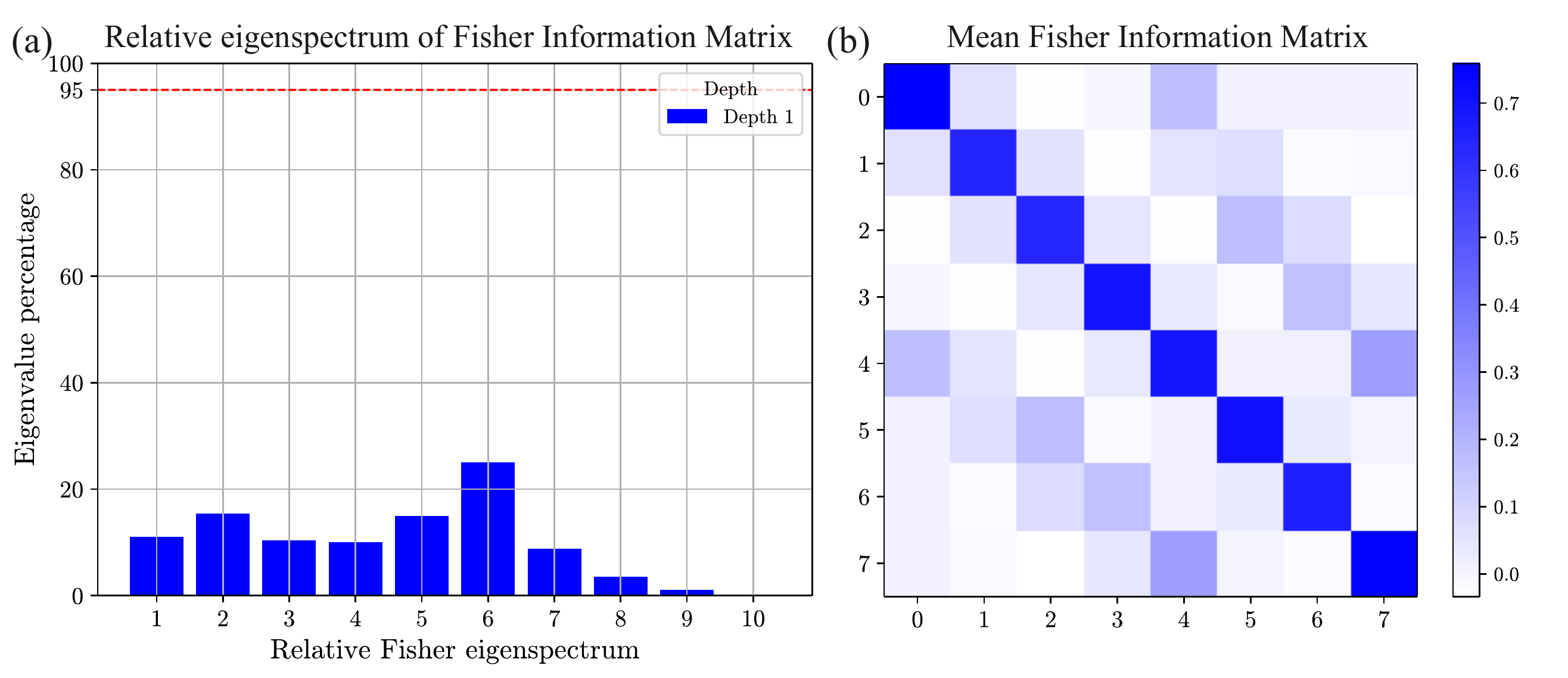}
    \caption{(\textbf{{a}}) Normalised histogram of the Fisher eigenvalue spectrum. The
    small number of close-to-zero eigenvalues indicates resilience to the
    barren-plateau problem \cite{abbas2021power}. (\textbf{{b}}) Averaged normalised
    Fisher Information Matrix. The diagonal pattern indicates that the
    quantum circuit distributes gradients evenly across trainable
    parameters; weak anti-diagonal elements imply parameters are
    approximately independent.}
    \label{fig:fisher}
\end{figure*}

Following \cite{abbas2021power,Araz2022Fisher}, we compute the FIM of the
QDI circuit on a Gaussian dataset $x \sim \mathcal{N}(0, 1)$. By averaging
over $x$ and $y$, one obtains the mean FIM.
Figure~\ref{fig:fisher}a shows the normalised histogram of FIM
eigenvalues; the eigenvalues are well-distributed with no significant
concentration near zero. In Figure~\ref{fig:fisher}b, the diagonal structure
in the averaged FIM indicates that gradients are approximately evenly
allocated across parameters, while minimal off-diagonal elements suggest
weak cross-parameter entanglement and straightforward optimisation.

These results indicate that, on this synthetic input distribution, the isolated
QDI circuit shows no signature of barren plateaus, a necessary (though not
sufficient) condition for trainability on the RUL prediction task.

The Gaussian dataset above characterises the QDI architecture in isolation.
Because each QDI sits inside the QLSTM cell, however, its actual input is the
four-dimensional gate input projection $a_t^{\gamma}$ of
Equation~\eqref{eq:preact}, not a raw sensor reading or a standard normal. To test
whether the trainability indication survives on realistic inputs, we recomputed
the FIM on the empirical distribution of $a_t^{\gamma}$. Concretely, we
ran the ten trained HQRNNs over C-MAPSS windows and recorded, via forward hooks
on the twelve circuits (three QLSTM layers $\times$ four gates), the $360{,}000$
input vectors actually presented to the QDI layers; we then repeated the Fisher
analysis of Equation~\eqref{eq:fisher} drawing the inputs from this empirical
distribution instead of from $\mathcal{N}(0,1)$. The recorded inputs are centred
near zero with standard deviation $1.26$ and range $[-8.6, 11.1]$, a scale
comparable to $\mathcal{N}(0,1)$ but with noticeably heavier tails.

\begin{figure*}[ht!]
    \centering
    \includegraphics[width=1.8\columnwidth]{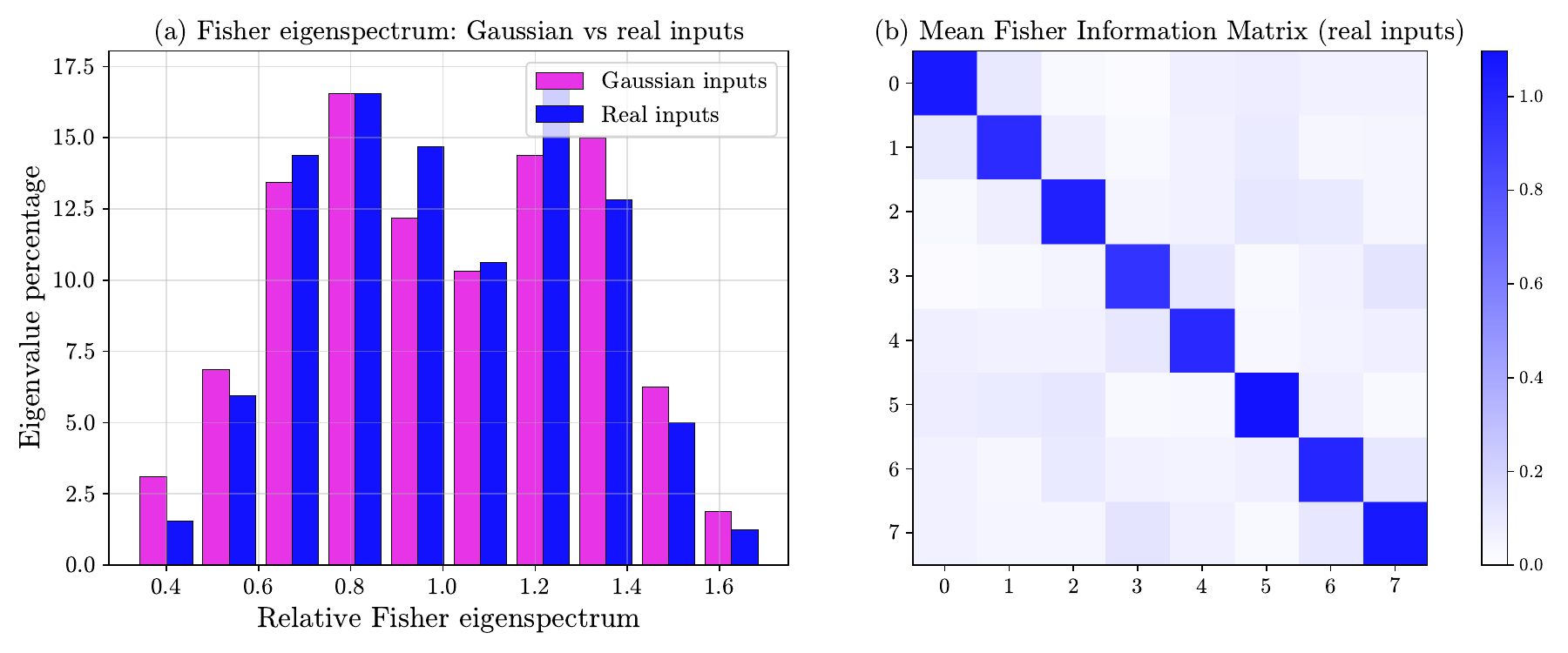}
    \caption{Fisher analysis on the empirical
    C-MAPSS-derived QDI input distribution. (\textbf{{a}}) FIM eigenvalue spectrum computed
    with the real circuit inputs overlaid on the synthetic-Gaussian baseline;
    the two distributions are nearly identical and neither
    places mass near zero. (\textbf{{b}}) Averaged normalised FIM for the real inputs,
    exhibiting the same diagonal structure as Figure~\ref{fig:fisher}b. Inputs
    were captured from the ten trained HQRNNs; the eigenvalue spectrum is
    unchanged when the trained weights are used in place of random ones.}
    \label{fig:fisher_real}
\end{figure*}

As shown in Figure~\ref{fig:fisher_real}, using the real inputs leaves the FIM
eigenvalue spectrum essentially unchanged: it overlaps the synthetic-Gaussian
spectrum almost exactly, the smallest eigenvalue remains $\approx 0.33$ with no
mass near zero, and the averaged FIM retains the diagonal structure of
Figure~\ref{fig:fisher}b. The same behaviour is observed when the random circuit
weights are replaced by the trained weights (a local FIM evaluated at the
learned solution). We therefore find no barren-plateau signature under the real
operating distribution, indicating that the trainability reading above is not an
artefact of the synthetic Gaussian assumption. As with the other circuit
diagnostics, this remains supporting rather than conclusive evidence and
characterises the isolated circuit.

\subsection{Expressivity: Fourier Series}\label{sec:fourier}

Quantum neural networks employing angle-based encoding can be interpreted
through the lens of truncated Fourier
series \cite{schuld2021effect,peters2023generalization,Parfait2024Fourier}.
The circuit's capacity to represent a function $f(\theta, x)$ can be
expressed as a multi-dimensional Fourier series whose degree of truncation
depends on the number of encoding repetitions \cite{perez2020data}. For two
encoded features, each repeated once, the function becomes
\begin{equation}
\begin{aligned}
f(\theta, x) &=
\braket{\psi(\theta, x)\mid M \mid \psi(\theta, x)} \\
&= \sum_{n = -1}^{1} \sum_{m = -1}^{1}
   c_{nm}(\theta)\, e^{i(nx_1 + m x_2)},
\end{aligned}
\end{equation}
where $|\psi(\theta, x)\rangle$ is the quantum state after all parameterised
operations, $M$ is the observable, and $c_{nm}$ are complex coefficients
determined by the circuit parameters. Although the maximal Fourier
frequencies are limited by circuit depth, they can be sufficient to capture
higher-order correlations in real-world datasets. A higher number of
non-zero coefficients represents more complex dependencies the model can
detect.

\begin{figure*}[ht!]
    \centering
    \includegraphics[width=0.62\textwidth]{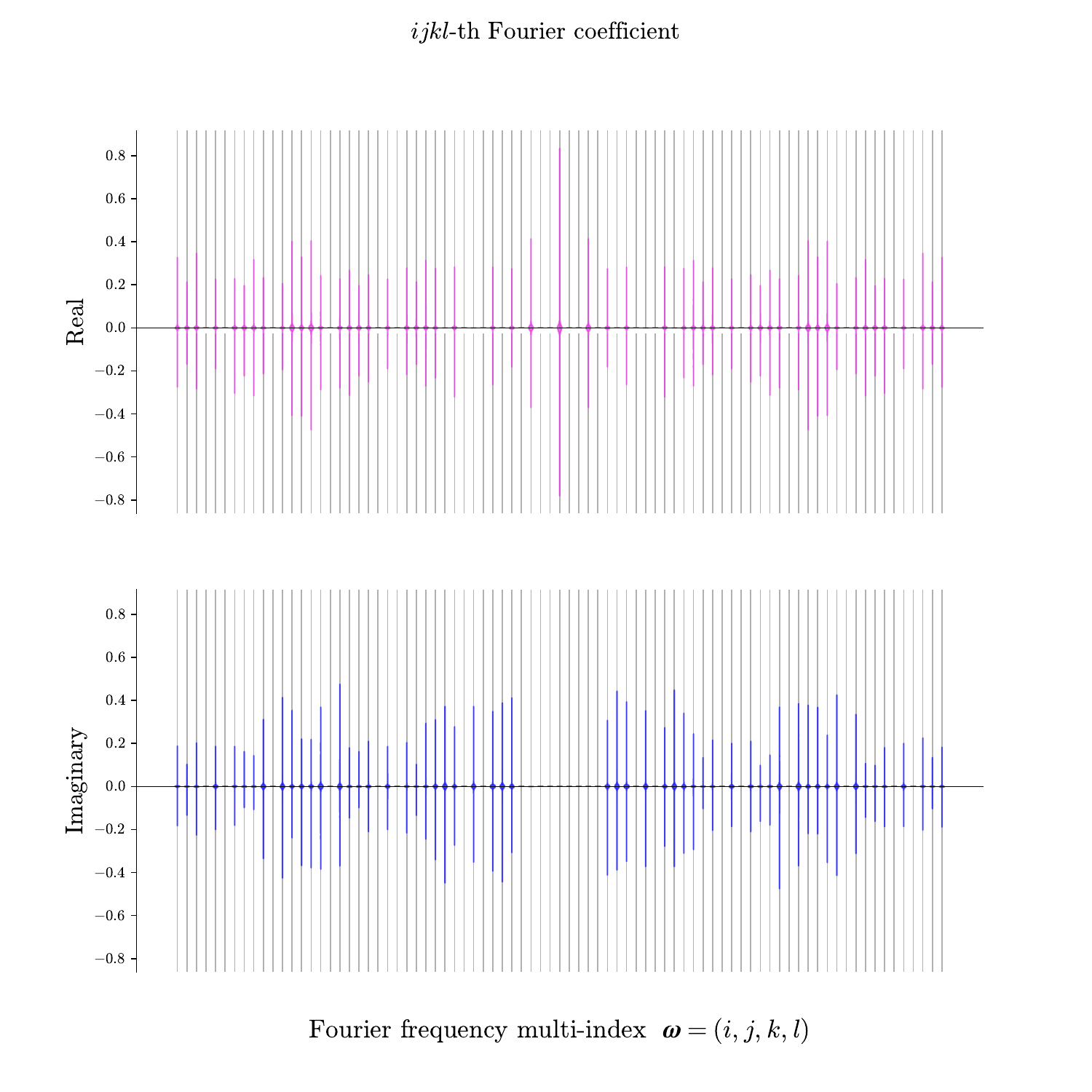}
    \caption{Fourier coefficients of a QDI layer with four
    input features. The horizontal axis enumerates the Fourier frequency
    multi-indices $\boldsymbol{\omega} = (i,j,k,l)$ accessible to the circuit
    (one vertical line per coefficient), and the two panels show, respectively,
    the real and imaginary parts of the corresponding complex coefficient
    $c_{\boldsymbol{\omega}}$; each coloured bar spans the range that coefficient
    takes over $1000$ random weight samples. The coefficients are amplitudes of
    the circuit's $[-1,1]$-valued expectation output and are therefore
    dimensionless, so the vertical axes carry no physical unit. The prevalence of
    non-zero coefficients ($\sim 67\%$) indicates substantial expressivity.}
    \label{fig:fourier}
\end{figure*}

In the QDI circuit we assess Fourier accessibility by encoding four
features, each appearing only once. Randomly initialising the circuit
parameters with $1000$ samples, we compute the real and imaginary parts of
the resulting Fourier spectrum, as shown in Figure~\ref{fig:fourier}. Out of
$161$ possible frequency components, $109$ have non-negligible amplitude
($\sim 67\%$), indicating that a large portion of the circuit's Fourier
space is accessible under random initialisation. The QDI layer therefore has
access to a broad Fourier spectrum, a property that can be advantageous when
representing complex high-dimensional signals.

\subsection{Summary of Circuit Analysis}\label{sec:circ_summary}

Taken together, the three analyses give a consistent picture of the QDI circuit.
First, the ZX-based rewriting shows that the circuit is compactly parameterised:
no parameter can be removed under the rules we applied. Second, the
Fisher-information analysis reveals no barren-plateau signature, neither on the
synthetic Gaussian inputs nor, importantly, on the empirical C-MAPSS input
distribution recorded from the trained HQRNNs (Section~\ref{sec:fisher}), which
is consistent with good optimisability. Third, the Fourier analysis shows that a
substantial portion of the circuit's Fourier spectrum is accessible under random
initialisation, indicating strong expressivity. In this way the three diagnostics
jointly characterise the QDI block as compact, optimisable, and expressive. We
stress, however, that they remain necessary rather than sufficient indicators:
even with the real-input Fisher result, they describe the circuit in
near-isolation and should be read as supporting rather than conclusive evidence
that the empirical gains reported in Section~\ref{sec:training} have a
specifically quantum-mechanical origin.

\section{Discussion}\label{sec:discussion}

The central empirical result of this study is that the HQRNN attains a lower test
error than its classical recurrent counterparts on C-MAPSS FD001 while using no
more trainable parameters. Crucially, the improvement holds against both
classical baseline families of Section~\ref{sec:training}: the RNN that reuses
the HQRNN's layer widths (and therefore carries more parameters) and the
RNN whose widths are reduced to match the HQRNN's parameter count.
Averaged over ten seeds, the HQRNN improves mean test RMSE and MAE by about
$5\%$ at an equal or smaller parameter budget. We read this as evidence of
parameter efficiency: replacing each gate's linear map with a Quantum
Depth-Infused circuit acts as a useful inductive bias, favouring smooth,
Fourier-rich mappings that resist overfitting when run-to-failure trajectories
are scarce, rather than as mere added capacity. The circuit-level analyses of
Section~\ref{sec:qcirc_analysis} are consistent with this reading: the QDI
circuit is compactly parameterised (ZX calculus), shows none of the near-zero
Fisher spectrum associated with barren plateaus, and can access a broad band of
frequencies (Fourier). As we stress below, however, these analyses are supporting
rather than conclusive evidence that the gain has a specifically
quantum-mechanical origin.

Set against this advantage is a concrete practical limitation. At present the QDI
circuits must be simulated classically, and that simulation is far slower and
more memory-intensive than running the matched classical RNN
(Section~\ref{sec:cost}): the gain is one of parameter efficiency and accuracy,
not of computational cost on today's hardware.

\subsection{Comparison with Prior Work}\label{sec:comparison}

In comparison with classical machine-learning models (Random Forest, LASSO
regression, etc.) and simple ANN models (MLP, CNN, LSTM) on this task, the
proposed HQRNN model shows the best performance in terms of RMSE
(Table~\ref{tab:comparison_1}).

However, to achieve state-of-the-art results, ensembles or combinations of
neural-network models with sophisticated feature preprocessing are
required (Table~\ref{tab:comparison_2}). For example, the ``Auto-RUL +
LSTM'' model \cite{asif2022deep} uses a refined degradation model to assign
RUL target labels, whereas the present work uses the simple piecewise-linear
degradation. It is therefore expected that the joint models in
Table~\ref{tab:comparison_2} outperform our raw HQRNN. This positioning
suggests that the proposed model is best used not only on its own but also
as a component within complex prognostics pipelines, e.g.,\ as a
quantum-enhanced recurrent block within transformer- or attention-based
joint architectures, to further improve their~performance.

\begin{table}[ht]
\caption{Comparison of the proposed HQRNN model with classical ML
    models and simple neural-network models on the C-MAPSS FD001 test set.
    Best RMSE score is in bold.}
\label{tab:comparison_1}
\centering
\begin{tabular}{llc}
    \toprule
    \textbf{{Type}}                          & \textbf{{Method}}                                & \textbf{{RMSE}}             \\
    \midrule
    \multirow{5}{*}{Classical ML} & RF \cite{zhang2016multiobjective}     & $17.91$          \\
                                  & LASSO \cite{zhang2016multiobjective}  & $19.74$          \\
                                  & SVM \cite{zhang2016multiobjective}    & $40.72$          \\
                                  & KNR \cite{zhang2016multiobjective}    & $20.46$          \\
                                  & GB \cite{zhang2016multiobjective}     & $15.67$          \\
    \midrule
    \multirow{3}{*}{ANN}          & MLP \cite{zhang2016multiobjective}    & $16.78$          \\
                                  & CNN \cite{sateesh2016deep}            & $18.45$          \\
                                  & LSTM \cite{zheng2017long}             & $16.14$          \\
    \midrule
    Proposed                      & HQRNN                                 & $\mathbf{15.46}$ \\
    \bottomrule
    \end{tabular}
\end{table}

\begin{table}[ht]
\caption{Comparison of the proposed HQRNN model with joint
    deep-learning models on the C-MAPSS FD001 test set. Best RMSE score is
    in bold.}
\label{tab:comparison_2}
\centering
\begin{tabular}{lc}
    \toprule
    \textbf{{Method}}                                          & \textbf{{RMSE}}             \\
    \midrule
    Transformer + TCNN \cite{wang2021remaining}     & $12.31$          \\
    CNN + LSTM \cite{kong2019convolution}           & $16.16$          \\
    LSTM + FCLCNN \cite{peng2021remaining}          & $11.17$          \\
    BLS + TCN \cite{yu2021prediction}               & $12.08$          \\
    Auto-RUL + LSTM \cite{asif2022deep}             & $\mathbf{7.78}$  \\
    \midrule
    HQRNN (proposed)                                & $15.46$          \\
    \bottomrule
    \end{tabular}
\end{table}

\subsection{Implications for Prognostics Pipelines}\label{sec:implications}

The empirical and analytical results have several practical implications for
prognostics and health management (PHM):

\begin{itemize}
    \item Parameter efficiency under data scarcity. The HQRNN matches
    or beats classical LSTMs with up to $\sim 2\times$ fewer parameters,
    which is attractive for fleets where only a small number of full
    run-to-failure trajectories are available.
    \item Component for hybrid pipelines. HQRNN can be inserted into
    existing prognostics pipelines as a drop-in replacement for an LSTM
    block, without requiring quantum hardware at inference: the QLSTM
    layers can be simulated on classical hardware for the small circuit
    sizes considered here, while still providing the inductive bias of a
    Fourier-rich feature extractor.
    \item Robustness to short observation windows. The ability to
    learn high-frequency components is particularly valuable when only
    short segments of sensor history are available, a common situation in
    operational aviation maintenance.
\end{itemize}

\subsection{Requirements and Roadmap for Noisy Quantum Hardware}\label{sec:nisq}

All results reported above were obtained with ideal, noiseless state-vector
simulation (PennyLane \texttt{lightning.qubit}); the QLSTM circuits were not
executed on physical hardware or under a device noise model. Because this scope
choice bears directly on how the present gains would translate to a quantum
processor, we make the hardware requirements, the expected noise level, and a
realistic development path explicit in this subsection. A quantitative
evaluation on noisy intermediate-scale quantum (NISQ)
hardware~\cite{Preskill2018quantumcomputingin,Bharti2022nisq} remains the
subject of dedicated follow-up work; the estimates below are intended to
delimit that effort rather than to substitute for it.

Each QDI block acts on four qubits and consists of two basic-entangler layers
surrounding a single angle-embedding layer (Section~\ref{sec:arch}). Being compiled to
elementary gates amounts to twelve single-qubit rotations (four $R_x$ in
each of the two entangling layers plus four $R_z$ encoding rotations), eight
two-qubit CNOT gates (four per entangling layer), and a four-qubit Pauli-$Y$
readout, giving a two-qubit-gate depth of two. 
A single forward pass over one
length-$30$ window evaluates $360$ such circuits (Section~\ref{sec:cost}), but
each individual circuit is shallow.
To first order, the probability that a circuit runs without any error is
$\prod_g (1-\varepsilon_g)$, where the product runs over every gate and
measurement $g$ in the circuit and $\varepsilon_g$ is the error probability
(infidelity) of that operation. The corresponding total error,
$1-\prod_g(1-\varepsilon_g)\approx\sum_g \varepsilon_g$, therefore grows roughly
linearly with the number of operations; and because two-qubit gates and
measurements are typically one to two orders of magnitude noisier than
single-qubit gates, it is dominated by the eight CNOTs and the four-qubit
readout~\cite{Bharti2022nisq,Cai2023quantum}. Inserting representative present-day
superconducting error rates (single-qubit-gate error
$\varepsilon_1\!\sim\!10^{-4}$, two-qubit-gate error
$\varepsilon_2\!\sim\!10^{-3}$--$10^{-2}$, and per-qubit readout error
$\varepsilon_\mathrm{ro}\!\sim\!10^{-2}$), the eight CNOTs contribute
$8\varepsilon_2\approx 1$--$8\%$ and the four-qubit readout
$4\varepsilon_\mathrm{ro}\approx 4\%$, while the twelve single-qubit rotations
add well under $1\%$. The estimated total is thus of order $5$--$15\%$ per
circuit execution, the upper end corresponding to median rather than
best-in-class calibration. This is small enough that the shallow QDI circuit is a
plausible near-term hardware target, yet large enough that error mitigation would
be needed before the sampled expectation values could support quantitative RUL
predictions. The deliberate choice of only four qubits and unit depth
(Section~\ref{sec:arch}) is what keeps this budget, and hence the noise, low.
Moderate noise levels are, moreover, not necessarily detrimental: appropriately
characterised hardware noise can act as an implicit regulariser during
quantum-network training~\cite{kuzmin2025method}.
Turning the same budget around yields concrete hardware targets. Since the eight
CNOTs contribute $\approx 8\varepsilon_2$, keeping their aggregate error below
$\sim\!5\%$ requires $\varepsilon_2 \lesssim 5\%/8 \approx 6\times10^{-3}$, a
two-qubit-gate fidelity of about $99.4\%$ or better; keeping the four-qubit
readout error comparably low ($4\varepsilon_\mathrm{ro} \lesssim 4\%$) requires a
readout fidelity above $\sim\!99\%$. These targets are already met by the best
current hardware: leading superconducting and trapped-ion processors report
two-qubit-gate fidelities of $\approx\!99.9\%$ (Google's
Willow~\cite{Acharya2025qec}) and $\approx\!99.8\%$ (Quantinuum's trapped-ion
system~\cite{Moses2023racetrack}), with readout fidelities around $99.5\%$,
although median rather than best-in-class devices remain somewhat below them. A
circuit as shallow as ours is therefore well within reach once error
mitigation~\cite{Temme2017error,Cai2023quantum} is applied. As for
connectivity, the basic-entangler layers couple the four qubits in a CNOT ring
$0\!-\!1\!-\!2\!-\!3\!-\!0$. On a linear nearest-neighbour chain the three chain
links ($0\!-\!1$, $1\!-\!2$, $2\!-\!3$) are native and only the ring-closing
$3\!-\!0$ link needs an added SWAP; IBM's heavy-hexagonal
lattice~\cite{Chamberland2020topological}, whose qubits have at most three
neighbours and which contains no four-qubit loops, embeds the four qubits along a
path with the same single-SWAP overhead; and trapped-ion processors with
all-to-all connectivity~\cite{Moses2023racetrack} realise the ring directly. In
every case the connectivity demand of a four-qubit circuit is modest and well
within existing device~topologies.
On hardware each circuit returns sampled
 expectation values rather than
exact ones, with statistical error scaling as $1/\sqrt{N_\text{shots}}$. Because a
Pauli expectation value on $[-1,1]$ has variance at most one, resolving it to a
precision of $\sim\!10^{-2}$ needs $N_\text{shots}\sim(10^{-2})^{-2}=10^{4}$
shots per circuit; across the $360$ circuits of one window this is about
$3.6\times10^{6}$ measurements. This sampling cost is distinct from, and
additional to, the classical-simulation cost reported in
Table~\ref{tab:cost}, and error-mitigation schemes such as zero-noise
extrapolation or probabilistic error
cancellation~\cite{Temme2017error,Cai2023quantum} multiply it further in
exchange for reduced bias. Encoding strategies designed explicitly around
finite-shot statistics, such as shot-based quantum
encoding~\cite{kyriacou2026shot}, offer a complementary route to reducing this
measurement overhead.
We therefore envisage a three-stage path to hardware deployment. In the
near term, the shallow four-qubit circuits studied here are already
within reach of noisy superconducting and trapped-ion devices when combined with
error mitigation: utility-scale experiments have extracted accurate expectation
values from $\sim\!100$-qubit processors running circuits substantially larger
than ours~\cite{Kim2023evidence}. A concrete first step is a noise-aware
study (device-calibrated noise models with finite shots, followed by execution
on a small physical device), together with noise-robust training
strategies~\cite{Berberich2024training}. In the medium term, the first
below-threshold demonstrations of quantum error
correction~\cite{Acharya2025qec} point towards early logical qubits that would
relax the per-gate error ceiling. In the long term, fault-tolerant
execution would remove the depth and gate-count constraints altogether, enabling
deeper and wider QDI circuits. Establishing where along this path a genuine quantum
advantage emerges, and whether the parameter-efficiency gains reported here
survive realistic noise and finite sampling, is the central question we leave
for future work.

\subsection{Limitations and Future Work}\label{sec:limitations}

Several limitations should be noted. First, the model is evaluated only on
the FD001 subset of C-MAPSS; the more challenging FD002--FD004 subsets,
which include multiple operating conditions and fault modes, were not
considered and remain an important target for future work. Second, the
present analysis assumes ideal noiseless simulation of the quantum
circuit; quantitative evaluation on noisy intermediate-scale quantum (NISQ)
hardware is left for follow-up work. Third, the piecewise-linear
degradation model used here is widely adopted but coarse; integrating the
HQRNN with the auto-RUL labelling approach \cite{asif2022deep} or with
attention-aware joint architectures \cite{wang2021remaining,deng2024prediction}
is a natural next step that may close the gap to the joint-model
state-of-the-art.

Looking further ahead, several research directions arise from this study.
One is to integrate quantum modules into established forecasting algorithms
such as random forests or gradient boosting, where quantum layers can serve
as advanced feature encoders. Transformer-based approaches such as
TabPFN \cite{hollmann2022tabpfn} or
Chronos \cite{ansari2024chronoslearninglanguagetime} could likewise benefit
from quantum enhancements, potentially improving their ability to model
complex temporal dependencies. As quantum hardware evolves, adaptive
strategies that dynamically vary the circuit size or depth could manage the
trade-off between representational capacity and trainability.

\section{Conclusions}\label{sec:conclusion}

Reliable remaining-useful-life prediction from short, noisy, and scarce
run-to-failure records is a persistent obstacle in prognostics and health
management. This study shows that a hybrid quantum--classical recurrent model can
address this regime with a markedly smaller parameter budget than its purely
classical counterpart: our Hybrid Quantum Recurrent Neural Network, which
replaces the linear transformation in each LSTM gate with a Quantum
Depth-Infused circuit, matches or exceeds equally sized classical LSTM baselines,
improving average RMSE and MAE by about $5\%$ at a matched parameter count on the
NASA C-MAPSS FD001 benchmark.

This work makes four core contributions. First, it introduces a hybrid quantum
recurrent architecture in which a Quantum Depth-Infused circuit replaces the
linear transformation inside every LSTM gate, and then it applies it to turbofan-engine
RUL prediction. Second, it establishes a fair, matched-parameter benchmark on
C-MAPSS FD001 with statistical-significance testing across ten seeds, showing
that the HQRNN generalises better than classical recurrent baselines of equal or
larger size. Third, it characterises the quantum gate at the circuit level
(redundancy via ZX calculus, trainability via the Fisher information spectrum, and
expressivity via the Fourier decomposition) to explain why so compact a circuit
trains stably and generalises well. Fourth, it quantifies the computational cost
of the approach and sets out the gate-fidelity, connectivity, finite-shot, and
error-mitigation requirements for eventual deployment on noisy quantum hardware
rather than leaving hardware feasibility unspecified.

Three broader implications follow. First, the result reinforces
parameter efficiency as the practically relevant advantage of hybrid
quantum models in data-scarce settings, where the number of full degradation
trajectories, not compute, is the binding constraint. Second, because the QLSTM
is a drop-in replacement for a classical recurrent block and, at the small
circuit sizes used here, is efficiently simulable on classical hardware, it can
be incorporated as a component within the transformer- and attention-based
joint pipelines that define the current state of the art rather than competing
with them as a stand-alone predictor. Third, the accompanying circuit-level
analysis (ZX calculus, Fisher information, and Fourier expressivity) provides
supporting evidence that the chosen QDI circuit is compact, trainable, and
expressive, offering design guidance for quantum-enhanced sequence models beyond
this specific task.

We emphasise that advanced joint deep-learning architectures still outperform a
stand-alone HQRNN, and that the empirical gains reported here reflect
parameter efficiency rather than a demonstrated quantum-mechanical mechanism.
Establishing when and why hybrid recurrent modules help to extend the study
to the harder FD002--FD004 subsets and to noisy hardware and integrating the
HQRNN into joint prognostics pipelines are the natural next steps towards
practical quantum-enhanced predictive maintenance.

\section*{Author contributions}

{O.T.:} Conceptualisation, Methodology, Software,
Formal analysis, Investigation, Visualisation,
Writing---original draft.
{A.K.:} Methodology, Software,
Formal analysis, Validation, Writing---review and editing.
{A.S. (Arsenii Senokosov):} Software, Data curation, Investigation,
Validation.
{A.S. (Asel Sagingalieva):} Methodology, Supervision,
Writing---review and editing.
{A.M.:} Conceptualisation, Supervision, Project
administration, Writing---review and editing. All authors have read and agreed to the published version of the manuscript.

\section*{Funding}

This research received no external funding.

\section*{Data availability}

The NASA Commercial Modular Aero-Propulsion System Simulation (C-MAPSS)
dataset analysed in this study is publicly available from the NASA
Prognostics Center of Excellence Data Set
Repository (\url{https://www.nasa.gov/intelligent-systems-division/discovery-and-systems-health/pcoe/pcoe-data-set-repository/}, accessed on 9 August 2026).
The code used in this study is available from the corresponding author on
reasonable request.

\section*{Acknowledgments}

During the preparation of this manuscript, the authors used generative AI tools
solely for routine grammar and language editing of the prose. No generative AI
tools were used in the conception or design of the study, in the analysis or
interpretation of data, or in the generation of figures, tables, or results. The
authors have reviewed and edited all AI-assisted text and take full
responsibility for the content of the publication.

\section*{Conflicts of interest}

All authors are (or were, at the time the work was conducted) employed by Terra Quantum AG. 
 The  authors declare that they have no known competing financial interests or
personal relationships that could have appeared to influence the work
reported in this paper.


\begin{thebibliography}{999}

\bibitem[Si et~al.(2011)Si, Wang, Hu, and Zhou]{si2011remaining}
Si, X.S.; Wang, W.; Hu, C.H.; Zhou, D.H.
\newblock Remaining useful life estimation--a review on the statistical data
  driven approaches.
\newblock {\em Eur. J. Oper. Res.} {\bf 2011}, {\em
  213},~1--14.

\bibitem[Lee et~al.(2014)Lee, Wu, Zhao, Ghaffari, Liao, and
  Siegel]{lee2014prognostics}
Lee, J.; Wu, F.; Zhao, W.; Ghaffari, M.; Liao, L.; Siegel, D.
\newblock Prognostics and health management design for rotary machinery
  systems—{R}eviews, methodology and applications.
\newblock {\em Mech. Syst. Signal Process.} {\bf 2014}, {\em
  42},~314--334.

\bibitem[Berghout and Benbouzid(2022)]{berghout2022systematic}
Berghout, T.; Benbouzid, M.
\newblock A systematic guide for predicting remaining useful life with machine
  learning.
\newblock {\em Electronics} {\bf 2022}, {\em 11},~1125.

\bibitem[Zhang et~al.(2016)Zhang, Lim, Qin, and Tan]{zhang2016multiobjective}
Zhang, C.; Lim, P.; Qin, A.K.; Tan, K.C.
\newblock Multiobjective deep belief networks ensemble for remaining useful
  life estimation in prognostics.
\newblock {\em IEEE Trans. Neural Networks Learn. Syst.} {\bf
  2016}, {\em 28},~2306--2318.

\bibitem[Kang et~al.(2021)Kang, Catal, and Tekinerdogan]{kang2021remaining}
Kang, Z.; Catal, C.; Tekinerdogan, B.
\newblock Remaining useful life ({RUL}) prediction of equipment in production
  lines using artificial neural networks.
\newblock {\em Sensors} {\bf 2021}, {\em 21},~932.

\bibitem[Huang et~al.(2019)Huang, Huang, and Li]{huang2019bidirectional}
Huang, C.G.; Huang, H.Z.; Li, Y.F.
\newblock A bidirectional LSTM prognostics method under multiple operational
  conditions.
\newblock {\em IEEE Trans. Ind. Electron.} {\bf 2019}, {\em
  66},~8792--8802.

\bibitem[Ferreira and Gon{\c{c}}alves(2022)]{ferreira2022remaining}
Ferreira, C.; Gon{\c{c}}alves, G.
\newblock {Remaining Useful Life prediction and challenges: A literature review
  on the use of Machine Learning Methods}.
\newblock {\em J. Manuf. Syst.} {\bf 2022}, {\em
  63},~550--562.

\bibitem[Box et~al.(2015)Box, Jenkins, Reinsel, and Ljung]{box2015time}
Box, G.E.; Jenkins, G.M.; Reinsel, G.C.; Ljung, G.M.
\newblock {\em Time Series Analysis: Forecasting and Control}; John Wiley \&
  Sons:  Hoboken, NJ, USA, 
  2015.

\bibitem[Shumway et~al.(2000)Shumway, Stoffer, and Stoffer]{shumway2000time}
Shumway, R.H.; Stoffer, D.S.; Stoffer, D.S.
\newblock {\em Time Series Analysis and Its Applications};  Springer:  Berlin/Heidelberg, Germany, 
2000; Volume~3.

\bibitem[Zhang(2003)]{zhang2003time}
Zhang, G.P.
\newblock Time series forecasting using a hybrid ARIMA and neural network
  model.
\newblock {\em Neurocomputing} {\bf 2003}, {\em 50},~159--175.

\bibitem[Hochreiter and Schmidhuber(1997)]{hochreiter1997long}
Hochreiter, S.; Schmidhuber, J.
\newblock Long short-term memory.
\newblock {\em Neural Comput.} {\bf 1997}, {\em 9},~1735--1780.

\bibitem[Gers et~al.(2000)Gers, Schmidhuber, and Cummins]{gers2000learning}
Gers, F.A.; Schmidhuber, J.; Cummins, F.
\newblock Learning to forget: Continual prediction with LSTM.
\newblock {\em Neural Comput.} {\bf 2000}, {\em 12},~2451--2471.

\bibitem[Graves et~al.(2013)Graves, Mohamed, and Hinton]{graves2013speech}
Graves, A.; Mohamed, A.R.; Hinton, G.
\newblock Speech recognition with deep recurrent neural networks.
\newblock In \emph{Proceedings of the 2013 IEEE International Conference on
  Acoustics, Speech and Signal Processing}; IEEE: Piscataway, NJ, USA, 
  2013; pp. 6645--6649.

\bibitem[Sak et~al.(2014)Sak, Senior, and Beaufays]{sak2014long}
Sak, H.; Senior, A.W.; Beaufays, F. 
\newblock Long short-term memory recurrent neural network architectures for
  large scale acoustic modeling.
\newblock In \emph{Proceedings of the Interspeech 2014}, Singapore, 14--18 September 2014; 
ISCA: Baixas, France, 2014; pp. 338--342.

\bibitem[Al-Selwi et~al.(2024)Al-Selwi, Hassan, Jadid~Abdulkadir, Muneer,
  Sumiea, Alqushaibi, and Ragab]{LSTM2024Summary}
Al-Selwi, S.; Hassan, M.F.; Jadid~Abdulkadir, S.; Muneer, A.; Sumiea, E.;
  Alqushaibi, A.; Ragab, M.
\newblock {RNN-LSTM: From applications to modeling techniques and
  beyond—Systematic review}.
\newblock {\em J. King Saud Univ.-Comput. Inf. Sci.} {\bf 2024}, {\em 36},~102068.

\bibitem[Bishop(2006)]{Bishop2007Pattern}
Bishop, C.M.
\newblock {\em Pattern Recognition and Machine Learning}; Information Science
  and Statistics; Springer: New York, NY, USA,  2006.

\bibitem[Goodfellow et~al.(2016)Goodfellow, Bengio, Courville, and
  Bengio]{Goodfellow2016Deep}
Goodfellow, I.; Bengio, Y.; Courville, A.; Bengio, Y.
\newblock {\em Deep Learning}; MIT Press: Cambridge, MA, USA,  2016; Volume~1.

\bibitem[Nielsen and Chuang(2010)]{Nielsen2011Quantum}
Nielsen, M.A.; Chuang, I.L.
\newblock {\em Quantum Computation and Quantum Information}; Cambridge
  University Press: Cambridge, UK,  2010.

\bibitem[Biamonte et~al.(2017)Biamonte, Wittek, Pancotti, Rebentrost, Wiebe,
  and Lloyd]{biamonte2017quantum}
Biamonte, J.; Wittek, P.; Pancotti, N.; Rebentrost, P.; Wiebe, N.; Lloyd, S.
\newblock Quantum machine learning.
\newblock {\em Nature} {\bf 2017}, {\em 549},~195--202.

\bibitem[Montanaro(2016)]{montanaro2016quantum}
Montanaro, A.
\newblock Quantum algorithms: An overview.
\newblock {\em npj Quantum Inf.} {\bf 2016}, {\em 2},~15023.

\bibitem[Preskill(2018)]{Preskill2018quantumcomputingin}
Preskill, J.
\newblock Quantum computing in the {NISQ} era and beyond.
\newblock {\em Quantum} {\bf 2018}, {\em 2},~79.

\bibitem[Rebentrost et~al.(2014)Rebentrost, Mohseni, and
  Lloyd]{rebentrost2014quantum}
Rebentrost, P.; Mohseni, M.; Lloyd, S.
\newblock Quantum support vector machine for big data classification.
\newblock {\em Phys. Rev. Lett.} {\bf 2014}, {\em 113},~130503.

\bibitem[Ciliberto et~al.(2018)Ciliberto, Herbster, Ialongo, Pontil, Rocchetto,
  Severini, and Wossnig]{ciliberto2018quantum}
Ciliberto, C.; Herbster, M.; Ialongo, A.D.; Pontil, M.; Rocchetto, A.;
  Severini, S.; Wossnig, L.
\newblock Quantum machine learning: A~classical perspective.
\newblock {\em Proc. R. Soc. A Math. Phys. Eng. Sci.} {\bf 2018}, {\em 474},~20170551.

\bibitem[Schuld and Petruccione(2018)]{schuld2018supervised}
Schuld, M.; Petruccione, F.
\newblock {\em Supervised Learning with Quantum Computers}; Quantum Science and
  Technology; Springer: Cham, Switzerland, 2018.

\bibitem[Cao et~al.(2017)Cao, Guerreschi, and Aspuru-Guzik]{cao2017quantum}
Cao, Y.; Guerreschi, G.G.; Aspuru-Guzik, A.
\newblock Quantum neuron: An elementary building block for machine learning on
  quantum computers.
\newblock {\em arXiv} {\bf 2017}, arXiv:1711.11240.

\bibitem[Alharbi and Ahmad(2024)]{Alharbi2024QCNN}
Alharbi, M.; Ahmad, S.
\newblock Deep Revamped Quantum Convolutional Neural Network on Fashion MNIST
  Dataset.
\newblock {\em Data Metadata} {\bf 2024}, {\em 3},~358--368.

\bibitem[Havl{\'\i}{\v{c}}ek et~al.(2019)Havl{\'\i}{\v{c}}ek, C{\'o}rcoles,
  Temme, Harrow, Kandala, Chow, and Gambetta]{havlivcek2019supervised}
Havl{\'\i}{\v{c}}ek, V.; C{\'o}rcoles, A.D.; Temme, K.; Harrow, A.W.; Kandala,
  A.; Chow, J.M.; Gambetta, J.M.
\newblock Supervised learning with quantum-enhanced feature spaces.
\newblock {\em Nature} {\bf 2019}, {\em 567},~209--212.

\bibitem[Schuld and Killoran(2019)]{schuld2019quantum}
Schuld, M.; Killoran, N.
\newblock Quantum machine learning in feature Hilbert spaces.
\newblock {\em Phys. Rev. Lett.} {\bf 2019}, {\em 122},~040504.

\bibitem[Emmanoulopoulos and
  Dimoska(2022)]{emmanoulopoulos2022quantummachinelearningfinance}
Emmanoulopoulos, D.; Dimoska, S.
\newblock Quantum machine learning in finance: Time series forecasting.
\newblock {\em arXiv} {\bf 2022}, arXiv:2202.00599.

\bibitem[Sagingalieva et~al.(2025)Sagingalieva, Komornyik, Senokosov, Joshi,
  Mansell, Tsurkan, Pinto, Pflitsch, and
  Melnikov]{sagingalieva2025photovoltaic}
Sagingalieva, A.; Komornyik, S.; Senokosov, A.; Joshi, A.; Mansell, C.;
  Tsurkan, O.; Pinto, K.; Pflitsch, M.; Melnikov, A.
\newblock Photovoltaic power forecasting using quantum machine learning.
\newblock {\em Sol. Energy} {\bf 2025}, {\em 302},~114016.
\newblock {\url{https://doi.org/10.1016/j.solener.2025.114016}}.

\bibitem[Kordzanganeh et~al.(2023)Kordzanganeh, Sekatski, Fedichkin, and
  Melnikov]{kordzanganeh2023exponentially}
Kordzanganeh, M.; Sekatski, P.; Fedichkin, L.; Melnikov, A.
\newblock An exponentially-growing family of universal quantum circuits.
\newblock {\em Mach. Learn. Sci. Technol.} {\bf 2023}, {\em
  4},~035036.
\newblock {\url{https://doi.org/10.1088/2632-2153/ace757}}.

\bibitem[Arthur et~al.(2022)]{arthur2022hybridquantumclassicalneuralnetwork}
Arthur, D.
\newblock A hybrid quantum-classical neural network architecture for binary
  classification.
\newblock {\em arXiv} {\bf 2022}, arXiv:2201.01820.

\bibitem[Haboury et~al.(2024)Haboury, Kordzanganeh, Melnikov, and
  Sekatski]{haboury2024information}
Haboury, N.; Kordzanganeh, M.; Melnikov, A.; Sekatski, P.
\newblock Information plane and compression-gnostic feedback in quantum machine
  learning.
\newblock {\em arXiv} {\bf 2024}, arXiv:2411.02313.

\bibitem[Bischof et~al.(2025)Bischof, Teodoropol, F{\"u}chslin, and
  Stockinger]{Bischof2025Hybrid}
Bischof, L.; Teodoropol, S.; F{\"u}chslin, R.M.; Stockinger, K.
\newblock Hybrid quantum neural networks show strongly reduced need for free
  parameters in entity matching.
\newblock {\em Sci. Rep.} {\bf 2025}, {\em 15},~4318.

\bibitem[Sun et~al.(2025)Sun, Li, Xiang, Yuan, Hu, Hua, Jiang, Zhu, and
  Fu]{SUN2025130226}
Sun, Y.; Li, D.; Xiang, Q.; Yuan, Y.; Hu, Z.; Hua, X.; Jiang, Y.; Zhu, Y.; Fu,
  Y.
\newblock Scalable quantum convolutional neural network for image
  classification.
\newblock {\em Phys. A Stat. Mech. Its Appl.} {\bf
  2025}, {\em 657},~130226.

\bibitem[Patapovich et~al.(2025)Patapovich, Periyasamy, Kordzanganeh, and
  Melnikov]{patapovich2025superposed}
Patapovich, V.; Periyasamy, M.; Kordzanganeh, M.; Melnikov, A.
\newblock Superposed parameterised quantum circuits.
\newblock {\em arXiv} {\bf 2025}, arXiv:2506.08749.

\bibitem[Broughton et~al.(2020)Broughton, Verdon, McCourt, Martinez, Yoo,
  Isakov, Massey, Halavati, Niu, Zlokapa,
  et~al.]{broughton2021tensorflowquantumsoftwareframework}
Broughton, M.; Verdon, G.; McCourt, T.; Martinez, A.J.; Yoo, J.H.; Isakov,
  S.V.; Massey, P.; Halavati, R.; Niu, M.Y.; Zlokapa, A.;  et~al.
\newblock Tensorflow quantum: A software framework for quantum machine
  learning.
\newblock {\em arXiv} {\bf 2020}, arXiv:2003.02989.

\bibitem[Haboury et~al.(2023)Haboury, Kordzanganeh, Schmitt, Joshi, Tokarev,
  Abdallah, Kurkin, Kyriacou, and Melnikov]{haboury2023supervised}
Haboury, N.; Kordzanganeh, M.; Schmitt, S.; Joshi, A.; Tokarev, I.; Abdallah,
  L.; Kurkin, A.; Kyriacou, B.; Melnikov, A.
\newblock A~supervised hybrid quantum machine learning solution to the
  emergency escape routing problem.
\newblock {\em arXiv} {\bf 2023}, arXiv:2307.15682.

\bibitem[Sagingalieva et~al.(2025)Sagingalieva, Lusnig, Cavalli, and
  Melnikov]{sagingalieva2025hybrid}
Sagingalieva, A.; Lusnig, L.; Cavalli, F.; Melnikov, A.
\newblock Hybrid quantum neural networks for computer-aided sex diagnosis in
  forensic and physical anthropology.
\newblock {\em Informatics Med. Unlocked} {\bf 2025}, {\em 58},~101682.
\newblock {\url{https://doi.org/10.1016/j.imu.2025.101682}}.

\bibitem[Abbas et~al.(2021)Abbas, Sutter, Zoufal, Lucchi, Figalli, and
  Woerner]{abbas2021power}
Abbas, A.; Sutter, D.; Zoufal, C.; Lucchi, A.; Figalli, A.; Woerner, S.
\newblock The power of quantum neural networks.
\newblock {\em Nat. Comput. Sci.} {\bf 2021}, {\em 1},~403--409.

\bibitem[Berberich et~al.(2024)Berberich, Fink, Pranji{\'c}, Tutschku, and
  Holm]{Berberich2024training}
Berberich, J.; Fink, D.; Pranji{\'c}, D.; Tutschku, C.; Holm, C.
\newblock Training robust and generalizable quantum models.
\newblock {\em Phys. Rev. Res.} {\bf 2024}, {\em 6},~043326.

\bibitem[Saxena et~al.(2008)Saxena, Goebel, Simon, and Eklund]{cmapss}
Saxena, A.; Goebel, K.; Simon, D.; Eklund, N.
\newblock Damage propagation modeling for aircraft engine run-to-failure
  simulation.
\newblock In {\em 2008 International Conference on Prognostics and Health
  Management}; IEEE: Piscataway, NJ, USA, {2008}; pp. 1--9.

\bibitem[Saxena and Goebel(2008)]{saxena2008turbofan}
Saxena, A.; Goebel, K.
\newblock Turbofan engine degradation simulation data set.
\newblock {\em NASA Ames Progn. Data Repos.} {\bf 2008}, {\em
  18},~878--887.

\bibitem[Sateesh~Babu et~al.(2016)Sateesh~Babu, Zhao, and Li]{sateesh2016deep}
Sateesh~Babu, G.; Zhao, P.; Li, X.L.
\newblock Deep convolutional neural network based regression approach for
  estimation of remaining useful life.
\newblock In \emph{Proceedings of the Database Systems for Advanced Applications:
  21st International Conference, DASFAA 2016, Dallas, TX, USA,  16--19 April
  2016}; proceedings, part i 21; Springer:  Berlin/Heidelberg, Germany, 
  2016; pp. 214--228.

\bibitem[Zheng et~al.(2017)Zheng, Ristovski, Farahat, and Gupta]{zheng2017long}
Zheng, S.; Ristovski, K.; Farahat, A.; Gupta, C.
\newblock Long short-term memory network for remaining useful life estimation.
\newblock In \emph{Proceedings of the 2017 IEEE international conference on
  prognostics and health management (ICPHM)}; IEEE: Piscataway, NJ, USA,  2017; pp. 88--95.

\bibitem[Wang et~al.(2021)Wang, Cheng, and Song]{wang2021remaining}
Wang, H.K.; Cheng, Y.; Song, K.
\newblock Remaining useful life estimation of aircraft engines using a joint
  deep learning model based on TCNN and transformer.
\newblock {\em Comput. Intell. Neurosci.} {\bf 2021}, {\em
  2021},~5185938.

\bibitem[Deng and Zhou(2024)]{deng2024prediction}
Deng, S.; Zhou, J.
\newblock Prediction of remaining useful life of aero-engines based on
  {CNN-LSTM-Attention}.
\newblock {\em Int. J. Comput. Intell. Syst.}
  {\bf 2024}, {\em 17},~232.

\bibitem[Yu et~al.(2021)Yu, Wang, and Li]{yu2021prediction}
Yu, K.; Wang, D.; Li, H.
\newblock A prediction model for remaining useful life of turbofan engines by
  fusing broad learning system and temporal convolutional network.
\newblock In \emph{Proceedings of the 2021 8th International Conference on
  Information, Cybernetics, and Computational Social Systems (ICCSS)}; IEEE: Piscataway, NJ, USA,
  2021; pp. 137--142.

\bibitem[Peng et~al.(2021)Peng, Chen, Chen, Tang, Li, and
  Gui]{peng2021remaining}
Peng, C.; Chen, Y.; Chen, Q.; Tang, Z.; Li, L.; Gui, W.
\newblock A remaining useful life prognosis of turbofan engine using temporal
  and spatial feature fusion.
\newblock {\em Sensors} {\bf 2021}, {\em 21},~418.

\bibitem[Asif et~al.(2022)Asif, Haider, Naqvi, Zaki, Kwak, and
  Islam]{asif2022deep}
Asif, O.; Haider, S.A.; Naqvi, S.R.; Zaki, J.F.; Kwak, K.S.; Islam, S.R.
\newblock A deep learning model for remaining useful life prediction of
  aircraft turbofan engine on C-MAPSS dataset.
\newblock {\em IEEE Access} {\bf 2022}, {\em 10},~95425--95440.

\bibitem[Kurkin et~al.(2025)Kurkin, Hegemann, Kordzanganeh, and
  Melnikov]{kurkin2025forecasting}
Kurkin, A.; Hegemann, J.; Kordzanganeh, M.; Melnikov, A.
\newblock Forecasting steam mass flow in power plants using the parallel hybrid
  network.
\newblock {\em Eng. Appl. Artif. Intell.} {\bf 2025},
  {\em 160},~111912.
\newblock {\url{https://doi.org/10.1016/j.engappai.2025.111912}}.

\bibitem[Laskaris et~al.(2026)Laskaris, Morozov, Tarpanov, Seth, Procelewska,
  Sai~Gautam, Sagingalieva, Brasher, and Melnikov]{laskaris2026multi}
Laskaris, G.; Morozov, D.; Tarpanov, D.; Seth, A.; Procelewska, J.;
  Sai~Gautam, G.; Sagingalieva, A.; Brasher, R.; Melnikov, A.
\newblock Multi-objective optimization and quantum hybridization of
  equivariant deep learning interatomic potentials.
\newblock {\em Comput. Mater. Sci.} {\bf 2026}, {\em 270},~114742.
\newblock {\url{https://doi.org/10.1016/j.commatsci.2026.114742}}.

\bibitem[Lusnig et~al.(2024)Lusnig, Sagingalieva, Surmach, Protasevich, Michiu,
  McLoughlin, Mansell, de’Petris, Bonazza, Zanconati,
  et~al.]{lusnig2024hybrid}
Lusnig, L.; Sagingalieva, A.; Surmach, M.; Protasevich, T.; Michiu, O.;
  McLoughlin, J.; Mansell, C.; de’Petris, G.; Bonazza, D.; Zanconati, F.;
  et~al.
\newblock Hybrid quantum image classification and federated learning for
  hepatic steatosis diagnosis.
\newblock {\em Diagnostics} {\bf 2024}, {\em 14},~558.

\bibitem[Chen et~al.(2022)Chen, Yoo, and Fang]{chen2022quantum}
Chen, S.Y.C.; Yoo, S.; Fang, Y.L.L.
\newblock Quantum long short-term memory.
\newblock In \emph{Proceedings of the Icassp 2022-2022 IEEE International Conference
  on Acoustics, Speech and Signal Processing (ICASSP)}; IEEE: Piscataway, NJ, USA,  2022; pp.
  8622--8626.

\bibitem[Sagingalieva et~al.(2023)Sagingalieva, Kordzanganeh, Kenbayev,
  Kosichkina, Tomashuk, and Melnikov]{sagingalieva2023hybrid}
Sagingalieva, A.; Kordzanganeh, M.; Kenbayev, N.; Kosichkina, D.; Tomashuk, T.;
  Melnikov, A.
\newblock Hybrid quantum neural network for drug response prediction.
\newblock {\em Cancers} {\bf 2023}, {\em 15},~2705.

\bibitem[Anoshin et~al.(2024)Anoshin, Sagingalieva, Mansell, Zhiganov, Shete,
  Pflitsch, and Melnikov]{anoshin2024hybrid}
Anoshin, M.; Sagingalieva, A.; Mansell, C.; Zhiganov, D.; Shete, V.; Pflitsch,
  M.; Melnikov, A.
\newblock Hybrid quantum cycle generative adversarial network for small
  molecule generation.
\newblock {\em IEEE Trans. Quantum Eng.} {\bf 2024}, {\em
  5},~2500514.
\newblock {\url{https://doi.org/10.1109/TQE.2024.3414264}}.

\bibitem[Lopatkin et~al.(2026)Lopatkin, Sagingalieva, Lusnig, Protasevich,
  Behnke, and Melnikov]{lopatkin2026quantum}
Lopatkin, V.; Sagingalieva, A.; Lusnig, L.; Protasevich, T.; Behnke, B.;
  Melnikov, A.
\newblock Quantum hybrid feature selector.
\newblock {\em EPJ Quantum Technol.} {\bf 2026}, {\em 13},~49.
\newblock {\url{https://doi.org/10.1140/epjqt/s40507-026-00491-1}}.

\bibitem[Xu et~al.(2019)Xu, Zhang, and Xiao]{xu2019training}
Xu, Z.Q.J.; Zhang, Y.; Xiao, Y.
\newblock Training behavior of deep neural network in frequency domain.
\newblock In \emph{Proceedings of the Neural Information Processing: 26th
  International Conference, ICONIP 2019, Sydney, NSW, Australia, 
  12--15 December 2019}; Proceedings, Part I 26; Springer:  Berlin/Heidelberg, Germany, 
  2019; pp. 264--274.

\bibitem[Schuld et~al.(2021)Schuld, Sweke, and Meyer]{schuld2021effect}
Schuld, M.; Sweke, R.; Meyer, J.J.
\newblock Effect of data encoding on the expressive power of variational
  quantum-machine-learning models.
\newblock {\em Phys. Rev. A} {\bf 2021}, {\em 103},~032430.

\bibitem[Caro et~al.(2022)Caro, Huang, Cerezo, Sharma, Sornborger, Cincio, and
  Coles]{caro2022generalization}
Caro, M.C.; Huang, H.Y.; Cerezo, M.; Sharma, K.; Sornborger, A.; Cincio, L.;
  Coles, P.J.
\newblock Generalization in quantum machine learning from few training data.
\newblock {\em Nat. Commun.} {\bf 2022}, {\em 13},~4919.

\bibitem[Coecke and Duncan(2011)]{coecke2011interacting}
Coecke, B.; Duncan, R.
\newblock Interacting quantum observables: Categorical algebra and
  diagrammatics.
\newblock {\em New J. Phys.} {\bf 2011}, {\em 13},~043016.

\bibitem[van~de Wetering(2020)]{van2020zx}
van~de Wetering, J.
\newblock {ZX}-calculus for the working quantum computer scientist.
\newblock {\em arXiv} {\bf 2020}, arXiv:2012.13966.

\bibitem[Peham et~al.(2022)Peham, Burgholzer, and Wille]{peham2022equivalence}
Peham, T.; Burgholzer, L.; Wille, R.
\newblock Equivalence checking of quantum circuits with the ZX-calculus.
\newblock {\em IEEE J. Emerg. Sel. Top. Circuits Syst.} {\bf 2022}, {\em 12},~662--675.

\bibitem[Wang et~al.(2024)Wang, Yeung, and Koch]{wang2024differentiating}
Wang, Q.; Yeung, R.; Koch, M.
\newblock Differentiating and integrating ZX diagrams with applications to
  quantum machine learning.
\newblock {\em Quantum} {\bf 2024}, {\em 8},~1491.

\bibitem[Amari(1998)]{amari1998gradient}
Amari, S.i.
\newblock Natural gradient works efficiently in learning.
\newblock {\em Neural Comput.} {\bf 1998}, {\em 10},~251--276.

\bibitem[Berezniuk et~al.(2020)Berezniuk, Figalli, Ghigliazza, and
  Musaelian]{berezniuk2020scale}
Berezniuk, O.; Figalli, A.; Ghigliazza, R.; Musaelian, K.
\newblock A scale-dependent notion of effective dimension.
\newblock {\em arXiv} {\bf 2020}, arXiv:2001.10872.

\bibitem[McClean et~al.(2018)McClean, Boixo, and
  Smelyanskiy]{mcclean2018barren}
McClean, J.R.; Boixo, S.; Smelyanskiy, V.N.
\newblock Barren plateaus in quantum neural network training landscapes.
\newblock {\em Nat. Commun.} {\bf 2018}, {\em 9},~4812.

\bibitem[Araz and Spannowsky(2022)]{Araz2022Fisher}
Araz, J.Y.; Spannowsky, M.
\newblock Classical versus quantum: Comparing tensor-network-based quantum
  circuits on {Large Hadron Collider} data.
\newblock {\em Phys. Rev. A} {\bf 2022}, {\em 106},~062423.

\bibitem[Peters and Schuld(2023)]{peters2023generalization}
Peters, E.; Schuld, M.
\newblock Generalization despite overfitting in quantum machine learning
  models.
\newblock {\em Quantum} {\bf 2023}, {\em 7},~1210.

\bibitem[Atchadé and Larson(2024)]{Parfait2024Fourier}
Atchadé, P.; Larson, K.
\newblock {Fourier Series Weight in Quantum Machine Learning}.
\newblock {\em Adv. Artif. Intell. Mach. Learn.} {\bf
  2024}, {\em 4},~1866--1890.

\bibitem[P{\'e}rez-Salinas et~al.(2020)P{\'e}rez-Salinas, Cervera-Lierta,
  Gil-Fuster, and Latorre]{perez2020data}
P{\'e}rez-Salinas, A.; Cervera-Lierta, A.; Gil-Fuster, E.; Latorre, J.I.
\newblock Data re-uploading for a universal quantum classifier.
\newblock {\em Quantum} {\bf 2020}, {\em 4},~226.

\bibitem[Kong et~al.(2019)Kong, Cui, Xia, and Lv]{kong2019convolution}
Kong, Z.; Cui, Y.; Xia, Z.; Lv, H.
\newblock Convolution and long short-term memory hybrid deep neural networks
  for remaining useful life prognostics.
\newblock {\em Appl. Sci.} {\bf 2019}, {\em 9},~4156.

\bibitem[Bharti et~al.(2022)Bharti, Cervera-Lierta, Kyaw, Haug, Alperin-Lea,
  Anand, Degroote, Heimonen, Kottmann, Menke, Mok, Sim, Kwek, and
  Aspuru-Guzik]{Bharti2022nisq}
Bharti, K.; Cervera-Lierta, A.; Kyaw, T.H.; Haug, T.; Alperin-Lea, S.; Anand,
  A.; Degroote, M.; Heimonen, H.; Kottmann, J.S.; Menke, T.;  et~al.
\newblock Noisy intermediate-scale quantum algorithms.
\newblock {\em Rev. Mod. Phys.} {\bf 2022}, {\em 94},~015004.
\newblock {\url{https://doi.org/10.1103/RevModPhys.94.015004}}.

\bibitem[Cai et~al.(2023)Cai, Babbush, Benjamin, Endo, Huggins, Li, McClean,
  and O'Brien]{Cai2023quantum}
Cai, Z.; Babbush, R.; Benjamin, S.C.; Endo, S.; Huggins, W.J.; Li, Y.; McClean,
  J.R.; O'Brien, T.E.
\newblock Quantum error mitigation.
\newblock {\em Rev. Mod. Phys.} {\bf 2023}, {\em 95},~045005.
\newblock {\url{https://doi.org/10.1103/RevModPhys.95.045005}}.

\bibitem[Kuzmin et~al.(2025)Kuzmin, Somogyi, Pankovets, and
  Melnikov]{kuzmin2025method}
Kuzmin, V.; Somogyi, W.; Pankovets, E.; Melnikov, A.
\newblock Method for noise-induced regularization in quantum neural networks.
\newblock {\em Adv. Quantum Technol.} {\bf 2025}, {\em 8},~e00603.
\newblock {\url{https://doi.org/10.1002/qute.202400603}}.

\bibitem[{Google Quantum AI and Collaborators} et~al.(2025){Google Quantum AI
  and Collaborators}, Acharya, Abanin, et~al.]{Acharya2025qec}
Google Quantum AI and Collaborators. 
\newblock Quantum error correction below the surface code threshold.
\newblock {\em Nature} {\bf 2025}, {\em 638},~920--926.
\newblock {\url{https://doi.org/10.1038/s41586-024-08449-y}}.

\bibitem[Moses et~al.(2023)Moses, Baldwin, Allman, Ancona, Ascarrunz, Barnes,
  Bartolotta, Bjork, Blanchard, Bohn, et~al.]{Moses2023racetrack}
Moses, S.A.; Baldwin, C.H.; Allman, M.S.; Ancona, R.; Ascarrunz, L.; Barnes,
  C.; Bartolotta, J.; Bjork, B.; Blanchard, P.; Bohn, M.;  et~al.
\newblock A race-track trapped-ion quantum processor.
\newblock {\em Phys. Rev. X} {\bf 2023}, {\em 13},~041052.
\newblock {\url{https://doi.org/10.1103/PhysRevX.13.041052}}.

\bibitem[Temme et~al.(2017)Temme, Bravyi, and Gambetta]{Temme2017error}
Temme, K.; Bravyi, S.; Gambetta, J.M.
\newblock Error mitigation for short-depth quantum circuits.
\newblock {\em Phys. Rev. Lett.} {\bf 2017}, {\em 119},~180509.
\newblock {\url{https://doi.org/10.1103/PhysRevLett.119.180509}}.

\bibitem[Chamberland et~al.(2020)Chamberland, Zhu, Yoder, Hertzberg, and
  Cross]{Chamberland2020topological}
Chamberland, C.; Zhu, G.; Yoder, T.J.; Hertzberg, J.B.; Cross, A.W.
\newblock Topological and subsystem codes on low-degree graphs with flag
  qubits.
\newblock {\em Phys. Rev. X} {\bf 2020}, {\em 10},~011022.
\newblock {\url{https://doi.org/10.1103/PhysRevX.10.011022}}.

\bibitem[Kyriacou et~al.(2026)Kyriacou, Patapovich, Periyasamy, and
  Melnikov]{kyriacou2026shot}
Kyriacou, B.; Patapovich, V.; Periyasamy, M.; Melnikov, A.
\newblock Shot-based quantum encoding: a data-loading paradigm for quantum
  neural networks.
\newblock {\em Adv. Comput.} {\bf 2026}, {\em 1},~e70004.
\newblock {\url{https://doi.org/10.1002/adco.70004}}.

\bibitem[Kim et~al.(2023)Kim, Eddins, Anand, Wei, van~den Berg, Rosenblatt,
  Nayfeh, Wu, Zaletel, Temme, and Kandala]{Kim2023evidence}
Kim, Y.; Eddins, A.; Anand, S.; Wei, K.X.; van~den Berg, E.; Rosenblatt, S.;
  Nayfeh, H.; Wu, Y.; Zaletel, M.; Temme, K.;  et~al.
\newblock Evidence for the utility of quantum computing before fault tolerance.
\newblock {\em Nature} {\bf 2023}, {\em 618},~500--505.
\newblock {\url{https://doi.org/10.1038/s41586-023-06096-3}}.

\bibitem[Hollmann et~al.(2023)Hollmann, M{\"u}ller, Eggensperger, and
  Hutter]{hollmann2022tabpfn}
Hollmann, N.; M{\"u}ller, S.; Eggensperger, K.; Hutter, F.
\newblock Tabpfn: {A} transformer that solves small tabular classification
  problems in a second.
\newblock In Proceedings of the International Conference on Learning
  Representations (ICLR),  Kigali, Rwanda,  1--5 May 2023.

\bibitem[Ansari et~al.(2024)Ansari, Stella, Turkmen, Zhang, Mercado, Shen,
  Shchur, Rangapuram, Arango, Kapoor,
  et~al.]{ansari2024chronoslearninglanguagetime}
Ansari, A.F.; Stella, L.; Turkmen, C.; Zhang, X.; Mercado, P.; Shen, H.;
  Shchur, O.; Rangapuram, S.S.; Arango, S.P.; Kapoor, S.;  et~al.
\newblock {Chronos: Learning the language of time series}.
\newblock {\em arXiv} {\bf 2024},  	arXiv:2403.07815.

\end{thebibliography}
\end{document}